\documentclass[]{openmoss}
\usepackage{helvet}

\usepackage{amsmath} 
\usepackage{natbib}
\usepackage{graphicx}
\usepackage{subcaption} 

\usepackage[toc,page,header]{appendix}
\usepackage[utf8]{inputenc} 
\usepackage[T1]{fontenc}    
\usepackage{hyperref}       
\usepackage{url}            
\usepackage{booktabs}       
\usepackage{lmodern}        
\usepackage{amsfonts}       
\usepackage{nicefrac}       
\usepackage{microtype}      
\usepackage{wrapfig}

\usepackage{amssymb}  
\usepackage{fontawesome}  
\usepackage{url}  

\usepackage{titletoc}

\usepackage{minitoc}

\usepackage{booktabs}
\usepackage{array}
\usepackage{etoolbox}

\definecolor{lightblue}{RGB}{200, 230, 255}  
\definecolor{headerblue}{RGB}{150, 200, 255} 

\usepackage{pgfplots}
\usepackage[utf8]{inputenc} 
\usepackage[T1]{fontenc}    
\usepackage{hyperref}       
\usepackage{url}            
\usepackage{booktabs}       
\usepackage{amsfonts}       
\usepackage{nicefrac}       
\usepackage{microtype}      
\usepackage{xcolor}         
\usepackage{graphicx}
\usepackage{float}
\usepackage{comment}
\usepackage{multirow} 
\usepackage{amsmath} 
\usepackage{makecell} 
\usepackage{siunitx}  
\usepackage{tikz}
\usepackage{pgf-pie} 
\usepackage{subcaption}
\usepackage{wrapfig}
\usepackage[export]{adjustbox}

\usepackage{ragged2e}      
\usepackage{tabularx}       
\usepackage{array}          
\usepackage{caption}        
\usepackage{enumitem}
\usepackage{pifont}
\usepackage[hang,flushmargin]{footmisc} 

\usepackage{tcolorbox}

\usepackage{tcolorbox}    
\tcbuselibrary{breakable}  
\tcbuselibrary{skins}      

\newtcolorbox{promptbox}[2][]{
    colback=white,         
    coltext=black,         
    arc=3mm,               
    boxrule=0.5pt,         
    colframe=black!60!white, 
    title={#2},              
    colbacktitle=black, 
    coltitle=white,        
    fonttitle=\bfseries, 
    top=8pt,               
    bottom=8pt,            
    left=10pt,             
    right=10pt,            
    breakable,             
    before upper={%
        \linespread{1}\selectfont 
        \setlength{\parskip}{1ex plus 0.2ex minus 0.2ex}
        \setlength{\parindent}{0pt}
    },
    #1                     
}

\usepackage{tabularx}
\usepackage{listings}

\definecolor{ForestGreen}{RGB}{34, 139, 34}
\definecolor{Red}{RGB}{255, 0, 0}

\definecolor{tickG}{rgb}{0.1, 0.588, 0.1}
\definecolor{crossR}{rgb}{0.588, 0.1, 0.1}
\newcommand{\cmark}{\textcolor{tickG}{\ding{52}}}
\newcommand{\xmark}{\textcolor{crossR}{\ding{56}}}

\definecolor{frenchblue}{rgb}{0.0, 0.45, 0.73}
\definecolor{babyblue}{rgb}{0.54, 0.81, 0.94}
\definecolor{classicrose}{rgb}{0.98, 0.8, 0.91}
\definecolor{beige}{rgb}{0.96, 0.96, 0.86}
\definecolor{forestgreen}{HTML}{2e7d43}

\definecolor{blue1}{HTML}{91BBE6}
\definecolor{blue2}{HTML}{3F90E0}
\definecolor{blue3}{HTML}{316FAD}

\definecolor{color1}{HTML}{FF9999}
\definecolor{color2}{HTML}{FF6666}
\definecolor{color3}{HTML}{FF3333}
\definecolor{color4}{HTML}{E60000}
\definecolor{color5}{HTML}{B30000}
\definecolor{color6}{HTML}{8CD98C}
\definecolor{color7}{HTML}{53c653}
\definecolor{color8}{HTML}{00B050}
\definecolor{color9}{HTML}{2d862d}
\definecolor{color10}{HTML}{206020}
\definecolor{color11}{HTML}{cca300}

\title{Thinking with Video: Video Generation as a Promising Multimodal Reasoning Paradigm}

\author{
    Jingqi Tong\textsuperscript{2,4,5,*}, 
    Yurong Mou\textsuperscript{2,4,5,*}, 
    Hangcheng Li\textsuperscript{2,4,5,*}, 
    Mingzhe Li\textsuperscript{2,4,5,*},
    Yongzhuo Yang\textsuperscript{4,5,*},\\
    Ming Zhang\textsuperscript{4}, 
    Qiguang Chen\textsuperscript{7}, 
    Tianyi Liang\textsuperscript{2,5},
    Xiaomeng Hu\textsuperscript{6},
    Yining Zheng\textsuperscript{1,3,5},\\
    Xinchi Chen\textsuperscript{1,3,4,5,$\dagger$}, 
    Jun Zhao\textsuperscript{4,$\dagger$},
    Xuanjing Huang\textsuperscript{1,3,4},
    Xipeng Qiu\textsuperscript{1,2,3,4,5,$\dagger$}
}

\affiliation[1]{\mbox{Institute of Trustworthy Embodied AI, Fudan University}}
\affiliation[2]{\mbox{Shanghai Innovation Institute}}
\affiliation[3]{\mbox{Shanghai Key Laboratory of Multimodal Embodied AI}}
\affiliation[4]{\mbox{College of Computer Science and Artificial Intelligence, Fudan University}}
\affiliation[5]{\mbox{OpenMOSS Team}}
\affiliation[6]{\mbox{The Chinese University of Hong Kong}}
\affiliation[7]{\mbox{Central South University}}

\abstract{
\begin{abstract}

The ``Thinking with Text'' and ``Thinking with Images'' paradigms significantly improve the reasoning abilities of large language models (LLMs) and vision-language models (VLMs). However, these paradigms have inherent limitations. (1) Images capture only single moments and fail to represent dynamic processes or continuous changes, and (2) The separation of text and vision as distinct modalities, which hinders unified multimodal understanding and generation. Therefore, we propose ``\textbf{Thinking with Video}'', a new paradigm that leverages video generation models such as \mbox{Sora-2} to use video frames as a unified medium for multimodal reasoning. To support this exploration, we developed the Video Thinking Benchmark (VideoThinkBench), which covers both vision-centric tasks (e.g., Eyeballing Puzzles) and text-centric tasks (e.g., GSM8K and MMMU). Our evaluation on VideoThinkBench establishes Sora-2 as a capable reasoner. On vision-centric tasks, Sora-2 is comparable to state-of-the-art (SOTA) VLMs, and even surpasses GPT-5 by 10\% on eyeballing puzzles. On text-centric tasks, Sora-2 achieves 92\% accuracy on MATH, and 69.2\% accuracy on MMMU. Furthermore, we systematically analyze the source of these abilities. We also find that self-consistency and in-context learning can improve Sora-2's performance. In summary, our findings show that the video generation model is the potential unified multimodal understanding and generation model, positioning ``Thinking with Video'' as a potential unified multimodal reasoning paradigm.
\end{abstract}

}

\contact{\email{jqtong25@m.fudan.edu.cn}, \email{\{xc\_chen, zhaoj19, xpqiu\}@fudan.edu.cn}}

\checkdata[Repository]{\url{https://github.com/tongjingqi/Thinking-with-Video}}
\checkdata[Benchmark]{\url{https://huggingface.co/datasets/OpenMOSS-Team/VideoThinkBench}}

\begin{document}

\maketitle
\renewcommand{\thefootnote}{}
\footnotetext{$^*$Equal contribution.\\$^\dagger$Corresponding authors.}
\renewcommand{\thefootnote}{\arabic{footnote}}


\vspace{-1.5em}
\section{Introduction}

\begin{figure*}[t]
    \centering 
    \includegraphics[width=\textwidth]{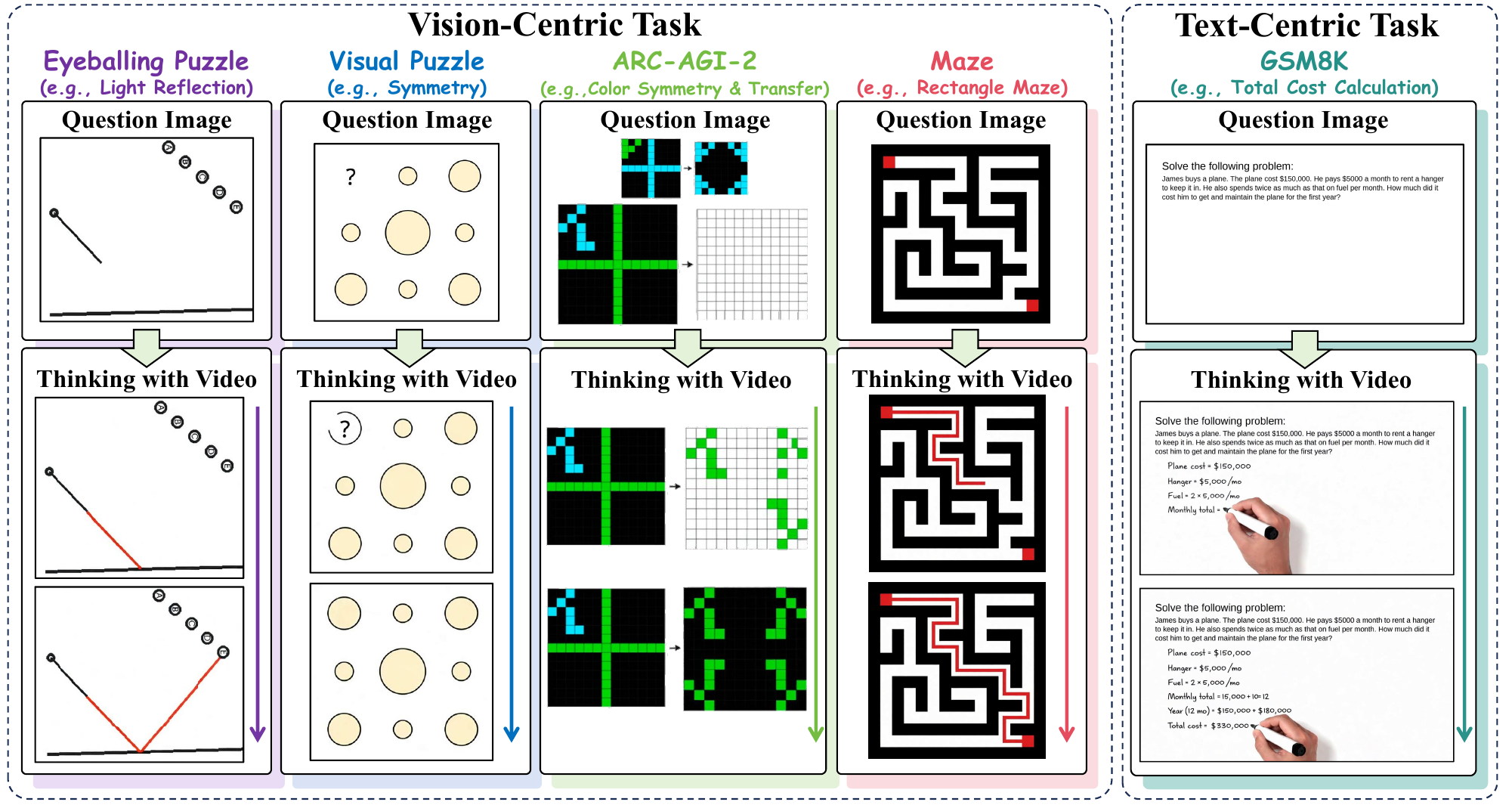}
    \caption{Examples of vision-centric and text-centric tasks in \textit{VideoThinkBench}, and Sora-2's ``Thinking with Video'' solutions. Vision-centric tasks are solved by reasoning about visual elements via \textbf{drawing and imagination}, including eyeballing puzzles, visual puzzles, ARC-AGI-2 and mazes. An example is shown for each. Typically, in the ``ray reflection'' problem from eyeballing puzzles, Sora-2 accurately draws the light path and finds the specific point it passes through. Text-centric tasks are solved by \textbf{text-based reasoning}, which are adapted from established benchmarks. A GSM8K example shows the model provides a written process and the answer in the video.}
    \label{fig:think_with_video_main}
\end{figure*}

Chain-of-Thought (CoT) significantly improves the reasoning ability of large language models (LLMs)~\citep{wei2022chain}, establishing ``Thinking with Text'' as a fundamental paradigm in AI reasoning.
OpenAI o3 can ``Think with Images'' in its Chain-of-Thought (CoT). ``Thinking with Images \citep{openai2024gpt4o,openai2025o3systemcard}'' is a paradigm that outputs images in CoT to help VLMs reason better. 
Models like Nano Banana~\citep{googledeepmind2025gemini} further demonstrate the capability of generating text embedded within images, bridging textual and visual reasoning.

Despite these advances, both ``Thinking with Text'' and ``Thinking with Images'' paradigms have inherent limitations.
(1) \textbf{Static constraints}: Images capture \textit{single moments} but cannot represent dynamic processes, temporal changes, or continuous transformations.
(2) \textbf{Modality separation}: Current approaches treat text and vision as \textit{separate modalities}, limiting the potential for unified multimodal understanding and generation. There is a lack of a unified framework that naturally integrates textual and visual reasoning within a coherent temporal structure.

Moving beyond the traditional paradigms of ``Thinking with Text'' and ``Thinking with Images'', we propose ``Thinking with Video''. It naturally enables human-like dynamic reasoning through video generation, such as \textit{drawing} and \textit{imagination}.
Video generation models, such as Sora-2, show great promise as unifying, general-purpose multimodal foundation models. By generating videos in the reasoning chain, models can:
(1) \textbf{Dynamic Reasoning}: Visualize dynamic processes (e.g., drawing lines to solve spatial puzzles) and represent continuous transformations.
(2) \textbf{Multimodal Fusion}: Embed text within video frames, as shown in Figure~\ref{fig:text-centric_eval_pipeline}, and achieve more natural alignment with human cognitive processes involving imagination and mental simulation.
Thus, ``Thinking with Video'' is potentially a unified multimodal reasoning paradigm.

To support this exploration, we developed the \textit{Video Thinking Benchmark (VideoThinkBench)}, which is designed to span a progression of core reasoning capabilities. VideoThinkBench encompasses vision-centric tasks and text-centric tasks. The vision-centric tasks reflect an increasing hierarchy of reasoning skills: from \textit{geometric intuition} (eyeballing puzzles), to \textit{visual pattern induction} (visual puzzles), to \textit{abstract rule induction} (ARC-AGI-2), and finally to \textit{spatial planning and search} (mazes). The text-centric tasks extend this progression into high-level \textit{language conceptual understanding and reasoning}. 



We evaluate on \textit{VideoThinkBench} and compare Sora-2 with SOTA VLMs, such as GPT-5~\citep{openai2025gpt5}, Claude Sonnet 4.5~\citep{anthropic2025claude4} and Gemini 2.5 Pro~\citep{comanici2025gemini}, as shown in Table~\ref{tab:summary_all_tasks}.
Furthermore, we systematically analyze the source of these abilities. The main findings of this work are as follows:

\begin{enumerate}

    \item On vision-centric tasks, Sora-2 is generally comparable to SOTA VLMs, and even \textbf{surpasses GPT-5 by 10\%} on eyeballing puzzles, demonstrating strong spatial reasoning and inductive abilities through drawing and imagination. For example, Sora-2 can paint lines to solve several spatial reasoning tasks. (Section~\ref{sec:eval_mmknowledge})


    \item On text-centric tasks, Sora-2 achieves surprising results. For text reasoning, Sora-2 achieves 98.9\% accuracy on GSM8K~\citep{cobbe2021gsm8k}, 92.0\% on MATH~\citep{hendrycks2021measuringmathematicalproblemsolving}, and 67.3\% on MMLU~\citep{hendrycks2020measuring}. For multimodal reasoning, Sora-2 achieves 75.7\% on MathVista~\citep{lu2023mathvista} and 69.2\% on MMMU~\citep{yue2024mmmu}. (Section~\ref{sec:text-centric_reasoning_task})


    \item Sora-2 is a \textbf{few-shot learner}. We evaluate it on ARC-AGI-2, which requires identifying patterns from input-output pairs and applying them to novel inputs. While SOTA VLMs struggle on ARC-AGI-2, we observe that Sora-2 can often make reasonable predictions but cannot strictly match the dataset annotations. (Section~\ref{sec:arc_agi_2}) Further experiments show that Sora-2 performs better when given more examples~\citep{dong2024survey}. (Section~\ref{sec:sora_few_shot})

    \item \textbf{Self-consistency} can improve Sora-2's performance in the verifiable video generation reasoning task. This reveals an underexplored direction: \textbf{test time scaling} in video generation reasoning tasks~\citep{wang2023selfconsistencyimproveschainthought}. (Section~\ref{sec:self_consistency})

    \item We systematically analyze the source of these abilities. Sora-2 maintains performance comparable to the original test set on adapted math problems. (Section~\ref{sec:exp_leakage})
    On text-centric tasks, Sora-2 struggles to generate a fully correct process. (Section~\ref{sec:text_reason_process})
    The source of Sora-2's text-centric reasoning abilities may originate from the prompt rewriter model. (Section~\ref{sec:source_text_reasoning_ability})
\end{enumerate}

In summary, our findings demonstrate that a video generation model is not only a general-purpose visual reasoning model, but also holds potential in unifying multimodal understanding and generation, positioning ``Thinking with Video'' as a potential unified multimodal reasoning paradigm.

\begin{table}[t]
    \centering
    \caption{Summary table of accuracy (\%) across all second-level tasks on the full test set of VideoThinkBench. For Sora-2: Eyeballing Puzzles uses Major Frame evaluation (see Section~\ref{sec:spatial_reasoning}), and text-centric tasks use audio evaluation results (see Section~\ref{sec:text-centric_eval}). Evaluation results of more models are shown in the Appendix (Tables~\ref{tab:minitest_vision_centric} and~\ref{tab:minitest_text_centric}).}
    \begin{tabular*}{\textwidth}{@{\extracolsep{\fill}}llcccc}
        \toprule
        \textbf{Category} & \textbf{Task} & \textbf{Sora-2} & \begin{tabular}{@{}c@{}}\textbf{Gemini} \\ \textbf{2.5 Pro}\end{tabular} & \begin{tabular}{@{}c@{}}\textbf{GPT-5} \\ \textbf{high}\end{tabular} & \begin{tabular}{@{}c@{}}\textbf{Claude} \\ \textbf{Sonnet 4.5}\end{tabular} \\
        \midrule
        \multirow{9}{*}{Vision-Centric} 
        & Eyeballing-Point & 44.7 & 27.8 & 33.6 & 36.2 \\
        & Eyeballing-Line & 38.0 & 21.0 & 24.0 & 26.3 \\
        & Eyeballing-Shape & 34.5 & 34.5 & 32.5 & 50.5 \\
        & Visual-Symmetry & 81.9 & 94.9 & 98.5 & 80.1 \\
        & Visual-Gradient & 51.9 & 83.7 & 66.7 & 69.9 \\
        & Visual-Comp. & 57.5 & 67.0 & 85.0 & 82.0 \\
        & ARC-AGI-2 & 1.3 & 1.9 & 0.5 & 5.3 \\
        & Maze & 13.3 & 0.0 & 0.0 & 0.0 \\
        & \textbf{Average} & \textbf{40.4} & \textbf{41.3} & \textbf{42.6} & \textbf{43.8} \\
        \midrule
        \multirow{5}{*}{Text-Centric}
        & Text-Only Math & 68.6 & 94.8 & 97.2 & 90.0 \\
        & Text-Only General Knowledge & 65.3 & 84.5 & 85.2 & 86.3 \\
        & Multimodal Math & 61.2 & 66.7 & 69.6 & 65.6 \\
        & Multimodal General Knowledge & 79.1 & 83.0 & 80.6 & 82.3 \\
        & \textbf{Average} & \textbf{68.6} & \textbf{82.3} & \textbf{83.2} & \textbf{81.1} \\
        \midrule
        \multicolumn{2}{l}{\textbf{Overall Average}} & \textbf{49.8} & \textbf{55.0} & \textbf{56.1} & \textbf{56.2} \\
        \bottomrule
    \end{tabular*}
    \label{tab:summary_all_tasks}
\end{table}

\section{VideoThinkBench and Evaluation}
\label{sec:eval}

\label{sec:benchmark_overview}

We introduce the \textit{Video Thinking Benchmark (VideoThinkBench)}, a comprehensive benchmark designed to evaluate the reasoning capabilities of video generation models through both vision-centric and text-centric tasks, as shown in~Figure~\ref{fig:think_with_video_main}.

\paragraph{Task Categories} 
Vision-centric tasks refer to tasks that are solved primarily through reasoning about visual elements via drawing and imagination. Text-centric tasks refer to tasks that are solved primarily through text-based reasoning processes.

\paragraph{Core Reasoning Abilities}
Across these two categories, \textit{VideoThinkBench} systematically evaluates a progression of five fundamental reasoning abilities: 





\begin{enumerate}
\item \textbf{Geometric intuition}: the basic ability to judge simple spatial relations (eyeballing puzzles).

\item \textbf{Visual pattern induction}: finding regularities in shapes, colors, or layouts (visual puzzles).

\item \textbf{Abstract rule induction}: discovering structured or rule-based transformations (ARC-AGI-2).

\item \textbf{Spatial planning and search}: the ability to plan multi-step actions (mazes).

\item \textbf{Language conceptual understanding and reasoning}: the ability required for high-level language, mathematical, and logical reasoning (text-centric tasks such as MATH and GSM8K).
\end{enumerate}

The first four ability types correspond to vision-centric tasks, while the last one represents text-centric tasks. While these abilities can overlap, this simple hierarchy provides a useful way to analyze what kinds of reasoning video-based models can perform.

\paragraph{Task Construction} 

Among the vision-centric tasks, eyeballing puzzles and mazes are designed by us, and visual puzzles are adapted from PuzzleVQA~\citep{chia2024puzzlevqa}. The original ARC-AGI-2~\citep{chollet2025arcagi2newchallengefrontier} is adapted for video generation. Vision-centric tasks are highly automatic. Samples of eyeballing puzzles, visual puzzles, and mazes can be generated in batches programmatically. Except for visual puzzles, all vision-centric tasks are \textbf{verifiable} for evaluation.

For the text-centric tasks, we select existing text-only reasoning benchmarks (e.g., MATH-500~\citep{cobbe2021training}, MMLU~\citep{wang2024mmlu}) and multimodal reasoning benchmarks (MathVista~\citep{lu2023mathvista}, MMMU~\citep{yue2024mmmu}), and sample a subset from most of the benchmarks for evaluation cost control. The problems are adapted for video generation. Specifically, we display the problem in the reference image input and prompt the model to show the written process in the video. The last frame and the audio of the generated video are evaluated.
\subsection{Vision-Centric Reasoning}
\label{sec:eval_mmknowledge}
Vision-centric tasks include eyeballing puzzles (Section~\ref{sec:spatial_reasoning}), visual puzzles (Section~\ref{sec:visual_puzzles}), ARC-AGI-2 (Section~\ref{sec:arc_agi_2}), mazes (Appendix~\ref{sec:maze}). Multiple reasoning abilities are needed to solve these tasks. For instance, solving an ARC-AGI-2 problem may involve using geometric intuition to recognize basic shapes, tracking changes, and applying induction to formulate an abstract rule that maps inputs to outputs.

\begin{figure*}[!t]
\centering 
\includegraphics[width=\textwidth]{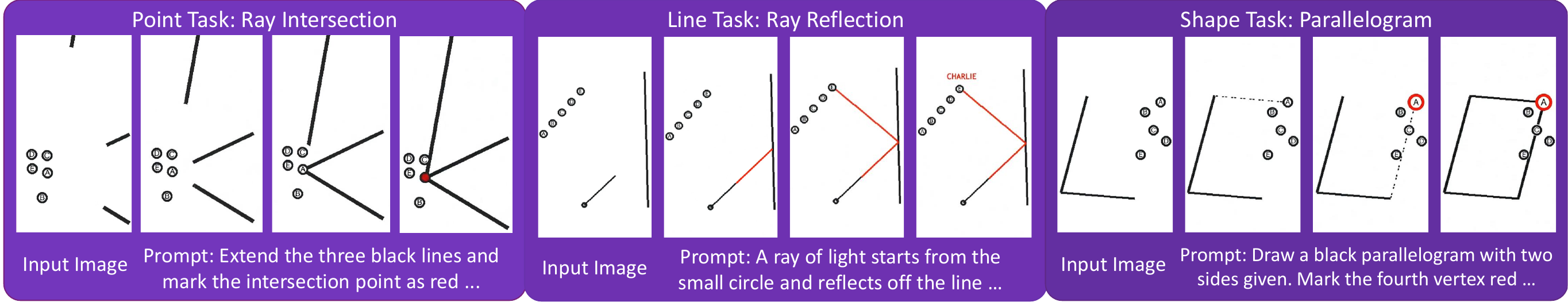}
\caption{\textbf{Three Examples of Sora-2 solving our custom benchmark of 21 eyeballing tasks and 1050 samples.} Each sample is a multiple choice question with image and text prompt, and is automatically evaluated. We divide tasks into three categories based on whether it constructs a point, a line or a shape. See Section~\ref{sec:spatial_reasoning} for details and Appendix~\ref{appendix_sec:eyeballing_prompts} for prompts. }
\label{fig:case_ray}
\end{figure*}

\begin{figure*}[!t]
\centering 
\includegraphics[width=\textwidth]{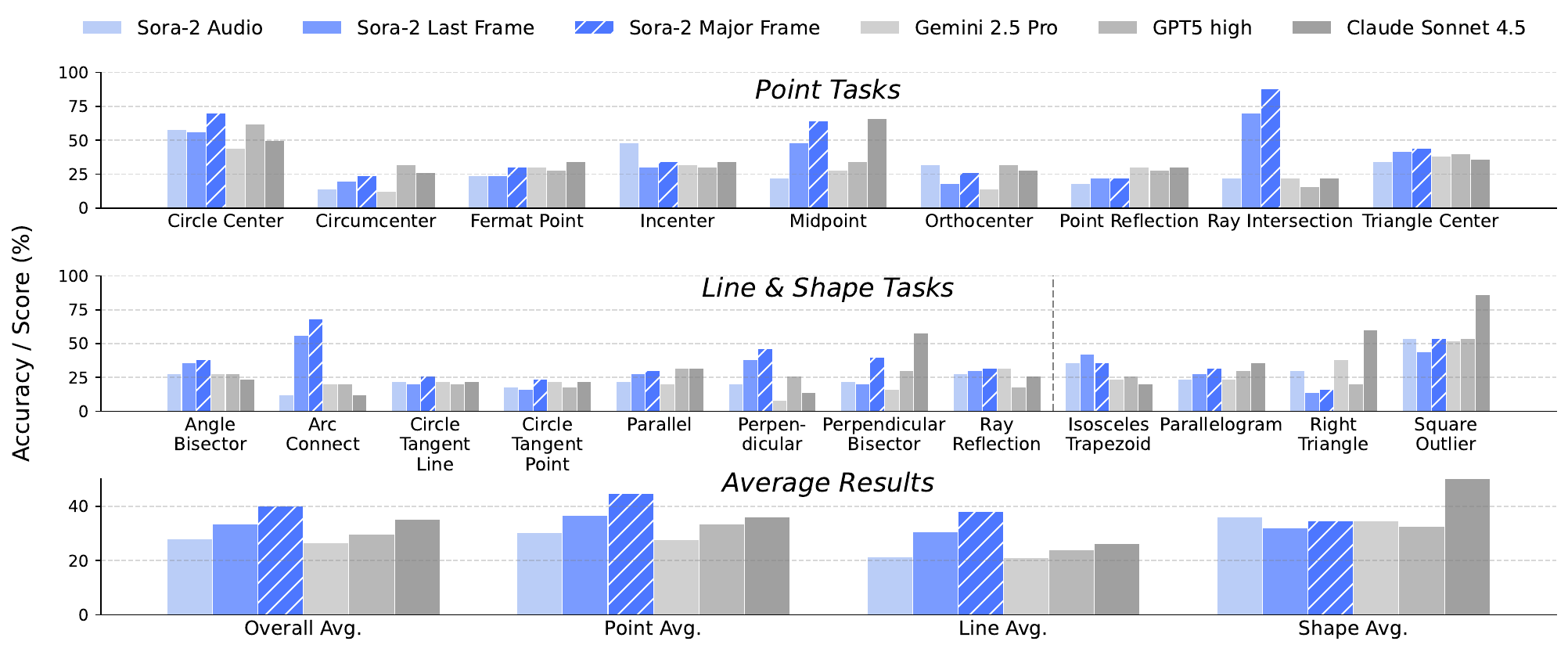}

\caption{Accuracy of Sora-2 using 3 evaluation methods and 3 VLMs on eyeballing tasks. For Sora-2, answers are derived from its generated audio (transcribed), the final video frame (which option is marked red), or by a majority vote across multiple frames. VLM answers are extracted from their text output. Table Version:~Table~\ref{tab:spatial_reasoning_results_table}. Details: Section~\ref{sec:spatial_reasoning}.}
\label{tab:spatial_reasoning_results}
\end{figure*}


\subsubsection{Geometric Reasoning: Eyeballing Puzzles}
\label{sec:spatial_reasoning}

\paragraph{Dataset and Setup} We designed a benchmark of 21 \textbf{verifiable} eyeballing puzzles (1,050 samples total) to test spatial reasoning, inspired by the ``eyeballing game''. As shown in Figure~\ref{fig:case_ray}, each multiple-choice puzzle tests a geometric concept (categorized as Point, Line, or Shape tasks). For Sora-2, we evaluated its video outputs via three methods: transcribing spoken answers from \textbf{Audio}, analyzing which option is marked red on the \textbf{Last Frame}, and taking a majority vote over sampled frames (\textbf{Major Frame}). Competing VLMs were prompted to output a text choice.


\paragraph{Results} Sora-2's video output proved most effective. The Major Frame Evaluation achieved the highest average accuracy of 40.2\%, outperforming Last Frame (33.4\%) and Audio (28.0\%) methods. This score surpasses all VLM competitors, including Claude 4.5 Sonnet (35.1\%), GPT-5 high (29.7\%), and Gemini 2.5 Pro (26.5\%) (Figure~\ref{tab:spatial_reasoning_results}).

\begin{tcolorbox}[colback=blue!10, colframe=blue!50, boxrule=1pt, arc=3mm, left=5pt, right=5pt, top=5pt, bottom=5pt,breakable]
\textbf{Takeaway 1}\quad Sora-2 generally \textbf{surpasses SOTA VLMs} on eyeballing puzzles, exhibiting geometric and physical reasoning abilities through drawing and imagination.
\end{tcolorbox}

\subsubsection{Inductive Reasoning: Visual Puzzles}
\label{sec:visual_puzzles}

\begin{figure*}[t]
\centering 
\includegraphics[width=\textwidth]{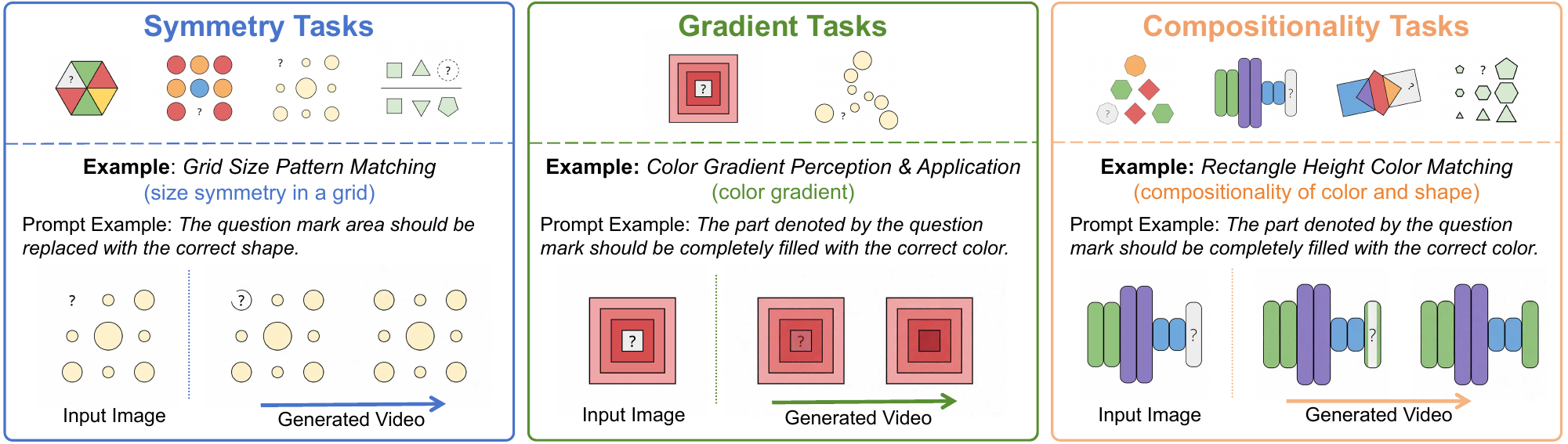}
\caption{Examples of the visual puzzles, selected and adapted from PuzzleVQA~\citep{chia2024puzzlevqa} to evaluate the ability of visual pattern induction. They are categorized into symmetry, gradient and compositionality tasks, with an example task shown for each. Sora-2 correctly solves the example task problems via video generation.}
\label{fig:visual_puzzles}
\end{figure*}

\begin{figure*}[t]
\centering 
\includegraphics[width=\textwidth]{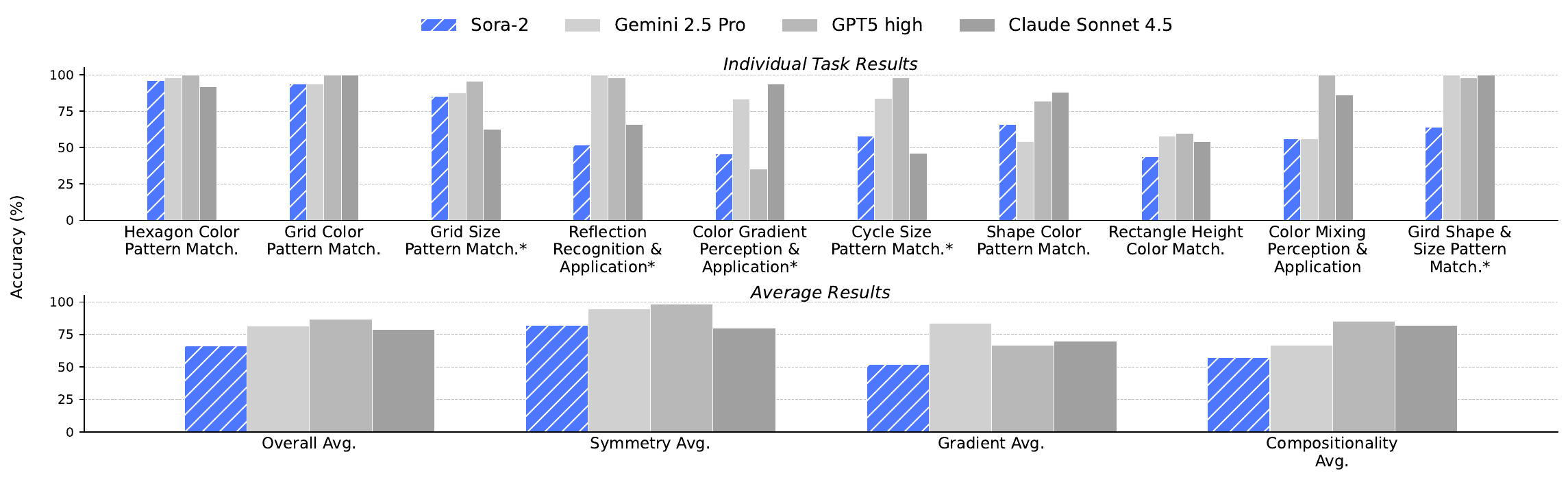}
\caption{Accuracy (\%) on the visual puzzle tasks. Table version:~Table~\ref{tab:visual_puzzle_result}. The ``*'' symbol represents that in the task, multiple-choice options are provided for the VLMs due to evaluation need, while Sora-2 is not given multiple-choice options across all the 10 tasks.}
\label{fig:visual_puzzles_bar}
\end{figure*}

\paragraph{Dataset and Setup} To evaluate \textbf{inductive reasoning}, we adapt ten visual puzzle types from PuzzleVQA~\citep{chia2024puzzlevqa} to video generation tasks, which are categorized into symmetry, gradient and compositionality tasks (Figure~\ref{fig:visual_puzzles}). Sora-2's output is manually evaluated on the ``best'' video frame (the frame with the lowest deviation from the solution, detailed in Appendix~\ref{app:visual_puzzle_diff}). VLMs are evaluated based on rules and given multiple-choice options in five of the tasks (Appendix~\ref{app:visual_puzzle_eval}).

\paragraph{Results} As shown in Figure~\ref{fig:visual_puzzles_bar}, Sora-2 demonstrates certain inductive reasoning capabilities across the visual puzzles. Furthermore, on the symmetry tasks, Sora-2 is competitive with Claude Sonnet 4.5. These results show that Sora-2 can recognize and apply patterns of color, shape, and size across symmetry, gradient, and compositionality tasks.



\begin{tcolorbox}[colback=blue!10, colframe=blue!50, boxrule=1pt, arc=3mm, left=5pt, right=5pt, top=5pt, bottom=5pt,breakable]
\textbf{Takeaway 2}\quad Sora-2 demonstrates inductive reasoning abilities in visual puzzles, including symmetry, gradient and compositionality tasks. Specifically, Sora-2's performance is comparable to that of Claude Sonnet 4.5 on symmetry tasks.
\end{tcolorbox}

\subsubsection{Few-Shot Learning: ARC-AGI-2}
\label{sec:arc_agi_2}


\paragraph{Dataset and Setup} We tested Sora-2 on the ARC-AGI-2 benchmark~\citep{chollet2025arcagi2newchallengefrontier}, which evaluates few-shot inductive reasoning over abstract grid transformations. The model must infer a transformation rule from examples and apply it to a test case. We evaluated Sora-2 by comparing the final video frame to the ground truth grid. We also manually analyzed 100 cases, categorizing them as Fully Correct, Mostly Correct, Partially Correct, or Wrong. See Figure~\ref{fig:case_arcagi2} for examples.

\paragraph{Results} Sora-2 achieved an accuracy of 1.3\%, close to strong VLMs like Gemini 2.5 Pro and GPT-5 high (Table~\ref{tab:arcagi_leaderboard_transposed}). Manual analysis (Table~\ref{tab:arcagi_manual}) of 100 samples revealed 3\% were ``Fully Correct'' and 14\% were ``Mostly Correct,'' suggesting the model often grasps the core rule but fails in execution. The example in Figure~\ref{fig:case_arcagi2} shows a case of self-correction during generation. A significant failure mode was the model not modifying the output area at all.

\begin{tcolorbox}[colback=blue!10, colframe=blue!50, boxrule=1pt, arc=3mm, left=5pt, right=5pt, top=5pt, bottom=5pt,breakable]
\textbf{Takeaway 3}\quad We find Sora-2 is a few-shot learner. While SOTA VLMs struggle on ARC-AGI-2 and achieve low accuracy, Sora-2 often makes reasonable predictions, though not strictly matching ground truths.
\end{tcolorbox}

\subsection{Text-Centric Reasoning Tasks}
\label{sec:text-centric_reasoning_task}

\begin{figure*}[t]
\centering 
\includegraphics[width=\textwidth]{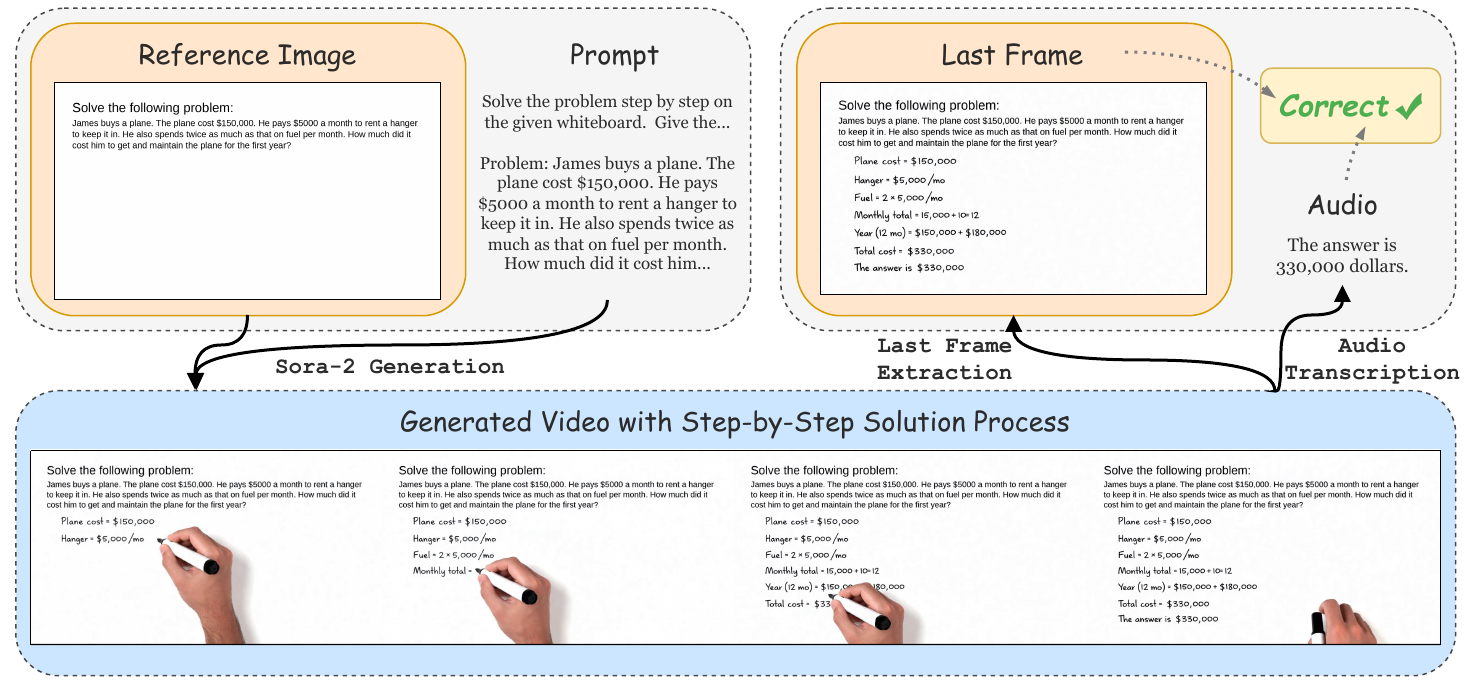}
\vspace{-20pt}
\caption{Input form and evaluation of text-centric tasks. The model accepts a text prompt and a reference image. The prompt contains the problem text and the reference image displays the entire problem. The model shows the textual solution process and the answer in the video, speaking the answer in the audio. We evaluate the answers from the video and audio independently. The last frame is extracted for video evaluation and the audio is transcribed for audio evaluation. For evaluation, we adopt an LLM-as-a-Judge approach, detailed in Section~\ref{sec:text-centric_eval}, and human alignment check is shown in Appendix~\ref{app:human_alignment_text}. Examples of testing multimodal reasoning problems are in Appendix~\ref{app:multimodal_reasoning_cases}.}
\label{fig:text-centric_eval_pipeline}
\end{figure*}


Language is expressed through visual symbols (e.g., written text), and humans naturally learn these symbols through \mbox{visual} perception~\citep{xing2025see,wei2025deepseek}. Since video generation models can learn complex visual patterns, they have the potential to learn language and reasoning in a visual form. This could help them solve math, logic, or abstract problems. Therefore, we are motivated to construct text-centric tasks to evaluate the model's language conceptual understanding and reasoning capabilities.

\subsubsection{Dataset Construction}
\label{sec:text-centric_data_construction}

\paragraph{Using Subsets of Selected Benchmarks} Text-centric tasks are adapted from established benchmarks, including text-only reasoning (e.g., GSM8K~\citep{cobbe2021gsm8k}, GPQA-diamond~\citep{rein2024gpqa}) and multimodal reasoning (e.g., MathVista~\citep{lu2023mathvista}, MMMU~\citep{yue2024mmmu}). For evaluation cost control, we sample a subset from most of the benchmarks, with detailed statistics listed in Appendix~\ref{app:bench_sample_distribution}.

\paragraph{Input and Output} As shown in Figure~\ref{fig:text-centric_eval_pipeline}, the input is a text prompt with the problem text and a reference image displaying the problem. The model generates a video of writing the solution, and an audio track that only states the final answer.

\subsubsection{Evaluation Setup}
\label{sec:text-centric_eval}

\paragraph{Video and Audio Evaluation} We evaluate video and audio answers independently, using the last frame for video evaluation and the transcription for audio evaluation.


\paragraph{LLM as a Judge} We adopt an LLM-as-a-Judge~\citep{zheng2023judging} approach for evaluation, using GPT-4o~\citep{openai2024gpt4o} as the judge. It is given the last frame or transcribed audio, as well as the question and the correct answer. Prompts and human alignment check are in Appendices~\ref{app:eval_prompt_text-centric} and~\ref{app:human_alignment_text}, respectively.

\subsubsection{Evaluation Results}
\label{sec:text_centric_eval_results}

Results of Sora-2 in Table~\ref{tab:text_centric_reasoning_statistics} show surprising performance on text-centric tasks. Its audio accuracy is comparable to SOTA VLMs on several benchmarks (e.g., GSM8K, MathVista), although gaps remain on more challenging ones (e.g., AIME, MMMU). Generally, Sora-2's audio accuracy is higher than its video accuracy, which may be due to difficulties in generating accurate written text, analyzed in Section~\ref{sec:text_reason_process}.

\begin{table}[t]
    \centering
    \large
    \setlength{\tabcolsep}{5pt} 
    \caption{Accuracy (\%) on subsets of text-only and multimodal reasoning benchmarks used for the text-centric tasks. The $\dagger$ symbol represents that the results are Avg@4. Sora-2 overall shows impressive reasoning capabilities, achieving performance comparable to SOTA VLMs on GSM8K, MATH-500, MathVista and MMBench in terms of audio accuracy, though noticeably lagging behind on more challenging datasets like AIME, GPQA, and MMMU.} 
    \resizebox{\textwidth}{!}{%
    \begin{tabular}{lccccc}
        \toprule
        {\textbf{Dataset}} & \textbf{Sora-2 Last Frame} & \textbf{Sora-2 Audio} & \textbf{Gemini 2.5 Pro} & \textbf{GPT-5 high} & \textbf{Claude Sonnet 4.5} \\
        \midrule
        \multicolumn{6}{c}{\textit{Text-Only Math Reasoning}} \\
        GSM8K & 75.7 & 98.9 & 98.9 & 100.0 & 100.0 \\
        MATH-500 & 67.0 & 92.0 & 99.0 & 99.0 & 98.0 \\
        AIME24$^\dagger$ & 38.3 & 46.7 & 93.3 & 95.0 & 75.0 \\
        AIME25$^\dagger$ & 33.3 & 36.7 & 88.0 & 94.6 & 87.0 \\
        \textbf{Average} & 53.6 & 68.6 & 94.8 & 97.2 & 90.0 \\
        \midrule
        \multicolumn{6}{c}{\textit{Text-Only General Knowledge Reasoning}} \\
        BBH & 69.8 & 80.6 & 90.0 & 94.6 & 93.8 \\
        MMLU & 69.1 & 67.3 & 87.7 & 86.0 & 89.5 \\
        MMLU-Pro & 72.0 & 76.5 & 87.1 & 91.4 & 95.7 \\
        GPQA & 51.5 & 57.6 & 86.4 & 85.7 & 83.4 \\
        SuperGPQA & 53.2 & 44.5 & 71.1 & 68.3 & 69.0 \\
        \textbf{Average} & 63.1 & 65.3 & 84.5 & 85.2 & 86.3 \\
        \midrule
        \multicolumn{6}{c}{\textit{Multimodal Math Reasoning}} \\
        MathVista & 67.6 & 75.7 & 70.0 & 67.5 & 72.5 \\
        MathVision & 44.9 & 46.7 & 63.3 & 71.6 & 58.7 \\
        \textbf{Average} & 56.3 & 61.2 & 66.7 & 69.6 & 65.6 \\
        \midrule
        \multicolumn{6}{c}{\textit{Multimodal General Knowledge Reasoning}} \\
        MMBench & 60.4 & 89.0 & 86.9 & 84.2 & 82.5 \\
        MMMU & 38.3 & 69.2 & 79.0 & 77.0 & 82.0 \\
        \textbf{Average} & 49.4 & 79.1 & 83.0 & 80.6 & 82.3 \\
        \midrule
        \textbf{Overall Average} & {57.0} & {67.8} & {84.7} & {85.8} & {83.7} \\
        \bottomrule
    \end{tabular}
    }
    \label{tab:text_centric_reasoning_statistics}
\end{table}

\begin{tcolorbox}[
    colback=blue!10,
    colframe=blue!50,
    boxrule=1pt,
    arc=3mm,
    left=5pt,
    right=5pt,
    top=5pt,
    bottom=5pt,breakable
]
\textbf{Takeaway 4}\quad On text-centric tasks, Sora-2 shows surprising performance on text and multimodal reasoning tasks. Sora-2 is able to embed text within video frames, enabling unified multimodal understanding and generation.
\end{tcolorbox}

\section{Analysis Experiment}
\label{sec:exp}
We conducted experiments on Sora-2 to analyze how to further enhance visual-centric reasoning capabilities and explore the origins of text-centric reasoning abilities.
\subsection{Enhancing Vision-Centric Reasoning Abilities}
We test the impact of the number of examples in few-shot learning scenarios to evaluate its in-context learning capabilities. We also experiment with self-consistency to evaluate the effect of test-time scaling on the model.

\subsubsection{More Examples Enhance In-Context Learning}
\label{sec:sora_few_shot}

Each sample in ARC-AGI-2 has multiple demonstration examples. To test Sora-2's few-shot learning capabilities, we retested Sora-2 on all 1000 training samples with only one example (1-shot) instead of all available examples. Since achieving a perfectly correct grid is difficult, we measure performance using ``pixel accuracy'': the percentage of pixels in the output area that match the ground truth. The results, comparing performance with all examples versus just one, are presented in Table~\ref{tab:pixel_accuracy_ranges}.


The results show a clear trend. Comparing to 1-shot, \textbf{few-shot} yields less low-accuracy (0 to 0.35) samples and more \textbf{high-accuracy} (0.65 to 1.0) samples. Sora-2's performance on this abstract reasoning task benefits from seeing multiple examples, confirming it is a few-shot learner.

\begin{tcolorbox}[
    colback=blue!10,
    colframe=blue!50,
    boxrule=1pt,
    arc=3mm,
    left=5pt,
    right=5pt,
    top=5pt,
    bottom=5pt,breakable
]

\textbf{Takeaway 5}\quad Sora-2 can achieve better in-context learning when more examples are provided. This is an underexplored direction for analyzing the in-context learning abilities of video generation models.
\end{tcolorbox}
\begin{table}[t]
    \centering

    \begin{minipage}[t]{0.48\textwidth}
        \centering
        \small
        \caption{In the few-shot setting of Sora-2, more samples fall within the accuracy range [0.65, 1], and fewer samples fall within the accuracy range [0, 0.35], compared to the one-shot setting. Few-shot uses all ARC-AGI-2 examples, while 1-shot uses only the first. Details: Section~\ref{sec:sora_few_shot}.}
        \resizebox{0.75\textwidth}{!}{%
        \begin{tabular}{lcc}
            \toprule
            \textbf{Accuracy Range} & \textbf{Few-Shot} & \textbf{1-Shot} \\
            \midrule
            0.00--0.35    & 743 & 788 \\
            0.35--0.65 & 127 & 117 \\
            0.65--1.00  & 130 & 95  \\
            \bottomrule
        \end{tabular}
        }
        \label{tab:pixel_accuracy_ranges}
    \end{minipage}
    \hfill%
    \begin{minipage}[t]{0.48\textwidth}
        \centering
        \small
        \caption{Sora-2's performance on the Arc Connect Puzzle by output modality. Vote Accuracy (5 Tries) means for each puzzle, we let Sora-2 generate 5 videos and choose the most common option as the final result. Details are in Appendix~\ref{sec:output_form}.}
        \resizebox{0.95\textwidth}{!}{%
        \begin{tabular}{lcc}
            \toprule
            \textbf{Evaluation Method} & \textbf{Single Try} & \textbf{Vote (5 Tries)} \\
            \midrule
            Audio         & 12\%                  & 12\%                      \\
            Last Frame    & 56\%                  & 66\%                      \\
            Major Frame         & \textbf{68\%}     & \textbf{90\%}                      \\
            \bottomrule
        \end{tabular}
        }
        \label{tab:output_modality_results}
    \end{minipage}%

\end{table}

\subsubsection{Self-Consistency Improves Performance}
\label{sec:self_consistency}

Self-consistency leverages the intuition that for a complex problem, multiple reasoning paths can lead to the same correct answer \citep{wang2023selfconsistencyimproveschainthought}. Arc Connect puzzle results in Table~\ref{tab:output_modality_results} demonstrate a similar result in video generation, shown by comparing the Last Frame and Major Frame evaluation methods.

Accuracy improves from 56\% for a single Last Frame analysis to 68\% for the Major Frame method. This is because the end of a video can be corrupted by SMPTE color bars or black screens, causing failures. By sampling across the video's duration, the Major Frame method acts as a denoising filter, capturing the model's most consistent belief.


This effect is magnified when using a majority vote over five retries. The Major Frame accuracy further leaps from 68\% to 90\%. This shows that test time scaling method of aggregating the result across multiple attempts can further improve its capabilities.

\begin{tcolorbox}[
    colback=blue!10,
    colframe=blue!50,
    boxrule=1pt,
    arc=3mm,
    left=5pt,
    right=5pt,
    top=5pt,
    bottom=5pt,breakable
]

\textbf{Takeaway 6}\quad We find that self-consistency can improve Sora-2’s performance in the verifiable video generation reasoning task. This reveals an
underexplored direction: test time scaling in video generation reasoning tasks.
\end{tcolorbox} 

\subsection{Analysis Experiment of Text-Centric Tasks}
We analyze the source of Sora-2's remarkable reasoning capabilities in text-centric tasks. We evaluate Sora-2 on adapted questions to eliminate the factor of test set leakage. Then, we analyze the Sora-2's text generation process. Finally, we evaluate Wan 2.5 for comparative analysis.

\subsubsection{Test Set Leakage Analysis}
\label{sec:exp_leakage}

\begin{wraptable}{R}{0.43\textwidth}
    \vspace{-1.0em} 
    \centering
    \small
    \setlength{\tabcolsep}{5pt}
    \caption{Sora-2’s accuracy on original and derived math reasoning problems with different numerical values. Performance remains consistent, thus excluding the risk of test data leakage and indicating Sora-2's inherent potential in text-centric tasks.}
    \begin{tabular}{lcc}
        \toprule
        \textbf{Dataset} & \textbf{Last Frame} & \textbf{Audio} \\
        \midrule
        GSM8K & 75.7 & 98.9 \\
        GSM8K (Derived) & 78.4 & 100.0 \\
        \midrule
        MATH-500 & 67.0 & 92.0 \\
        MATH-500 (Derived) & 75.0 & 91.0 \\
        \bottomrule
    \end{tabular}
    \label{tab:test_leakage_analysis}
    \vspace{-1em} 
\end{wraptable}

To investigate if Sora-2's performance in text-centric tasks comes from data leakage, we create new math reasoning problems. We use Qwen3-235B-A22B-Thinking-2507~\citep{yang2025qwen3} and Gemini 2.5 Pro~\citep{comanici2025gemini} to generate similar problems of GSM8K and MATH-500 problems, \mbox{respectively}. The prompts are in Appendix~\ref{app:data_leakage_prompt}. Each new problem shares the same solution structure as the original problem but uses different numerical values. Results (Table~\ref{tab:test_leakage_analysis}) show there is no significant performance difference compared to the original problems. This consistency suggests that Sora-2's performance comes from its inherent ability rather than test data leakage.

\subsubsection{Analysis of Sora-2's Reasoning Process}
\label{sec:text_reason_process}

To better understand Sora-2's text-centric reasoning, we sample 115 cases from the text-centric tasks that Sora-2 answered correctly in both video and audio. We manually analyze the written processes and categorize them as follows: (1) Completely Correct; (2) Logic Correct with Writing Errors; (3) Unreadable or Incorrect Logic; (4) Missing Solution Process; and (5) Process Unnecessary. Detailed definitions are provided in Appendix~\ref{app:solution_process_categories}.

\begin{figure}[t]
    \centering
    \begin{minipage}[t]{0.52\textwidth}
        \vspace{0pt}
        \centering
        \def\subfigwidth{0.48\linewidth} 
        \small
        \hfill
        \begin{subfigure}[t]{\subfigwidth}
            \centering
            \framebox{\includegraphics[width=\dimexpr\linewidth-2\fboxsep-2\fboxrule\relax]{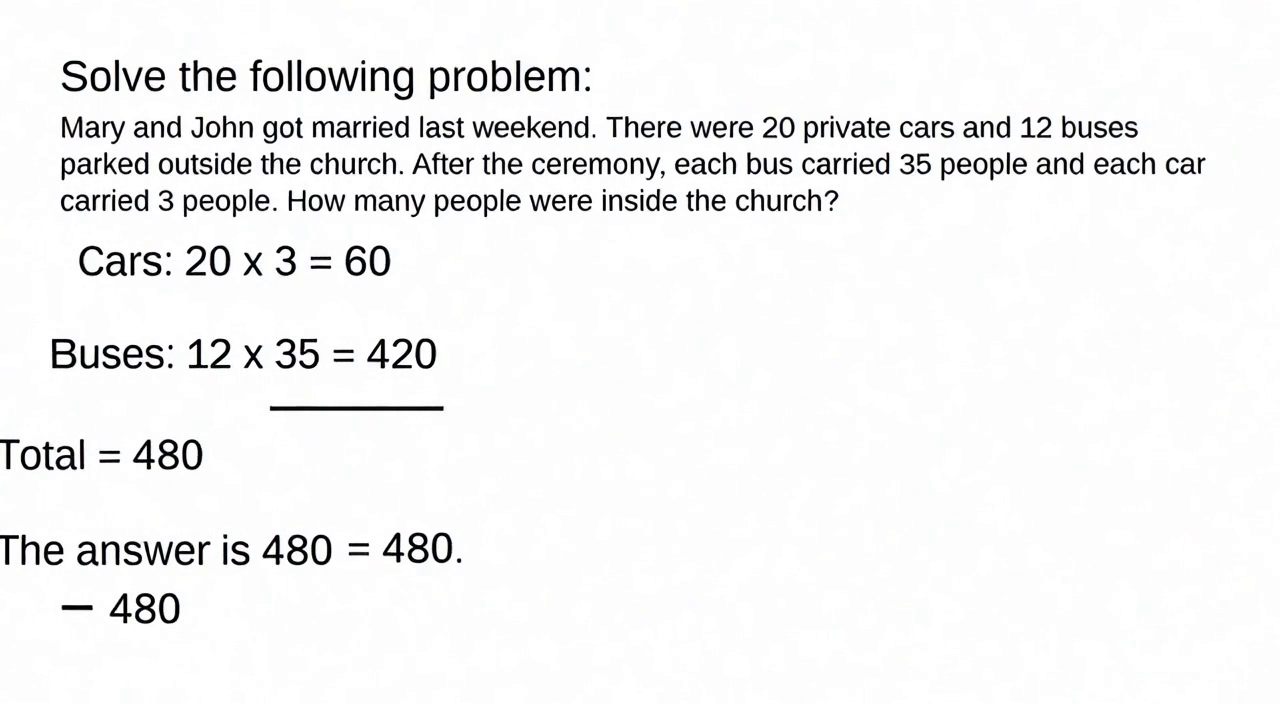}}
            \caption*{\makecell{Completely Correct \\ (13.91\%)}}
        \end{subfigure}%
        \hfill
        \begin{subfigure}[t]{\subfigwidth}
            \centering
            \framebox{\includegraphics[width=\dimexpr\linewidth-2\fboxsep-2\fboxrule\relax]{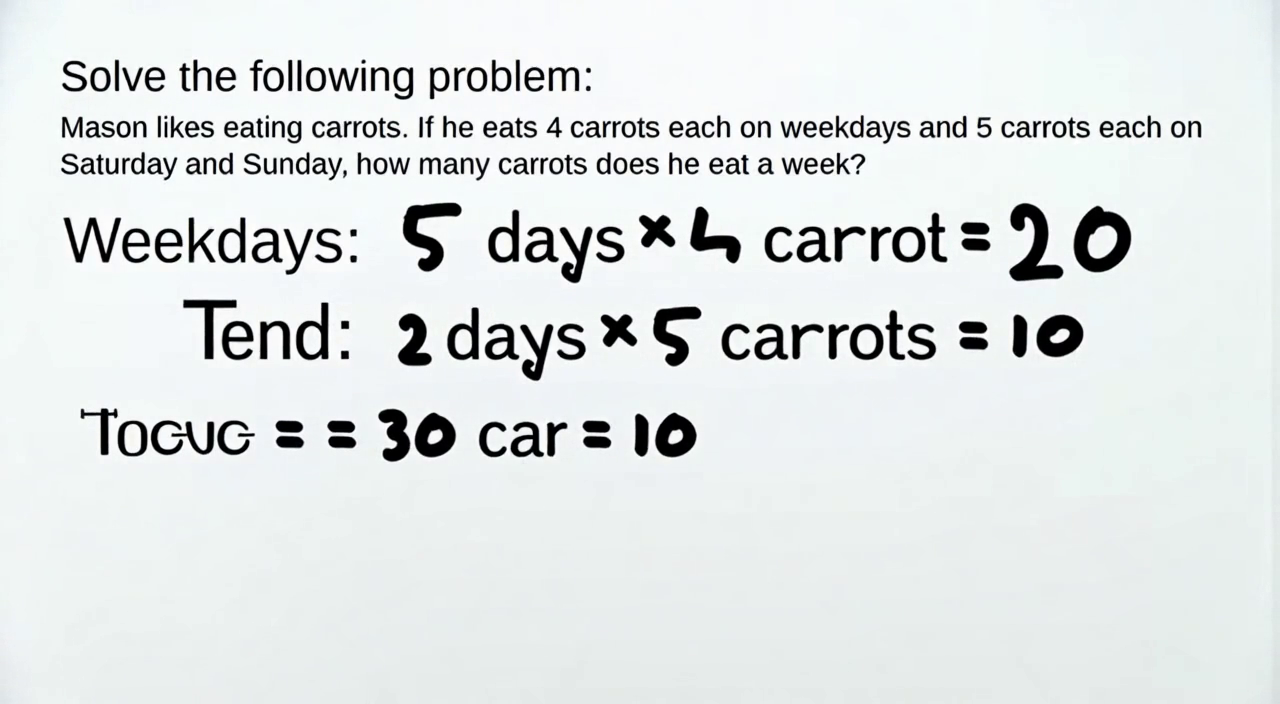}}
            \caption*{\makecell{Logic Correct with \\ Writing Errors (29.57\%)}}
        \end{subfigure}
        \vspace{1mm}
        
        \hfill
        \begin{subfigure}[t]{\subfigwidth}
            \centering
            \framebox{\includegraphics[width=\dimexpr\linewidth-2\fboxsep-2\fboxrule\relax]{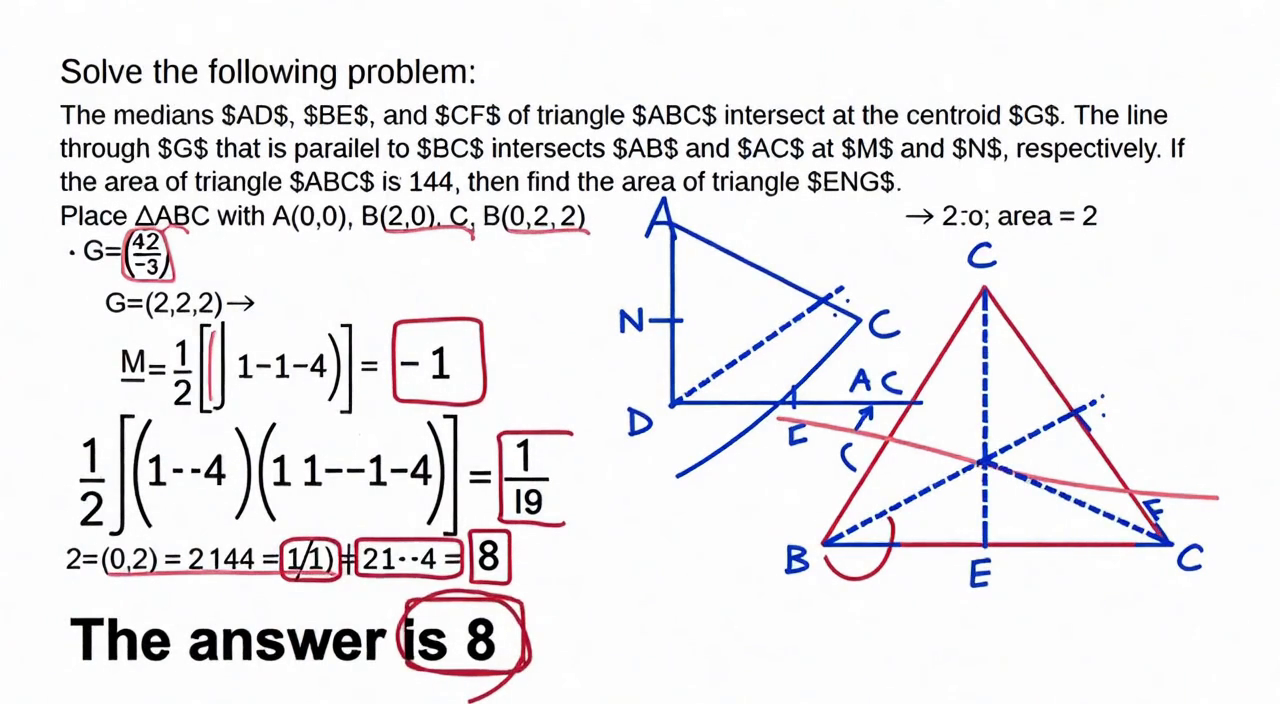}}
            \caption*{\makecell{Unreadable or Incorrect \\ Logic (43.48\%)}}
        \end{subfigure}%
        \hfill
       \begin{subfigure}[t]{\subfigwidth}
            \centering
            \framebox{\includegraphics[width=\dimexpr\linewidth-2\fboxsep-2\fboxrule\relax]{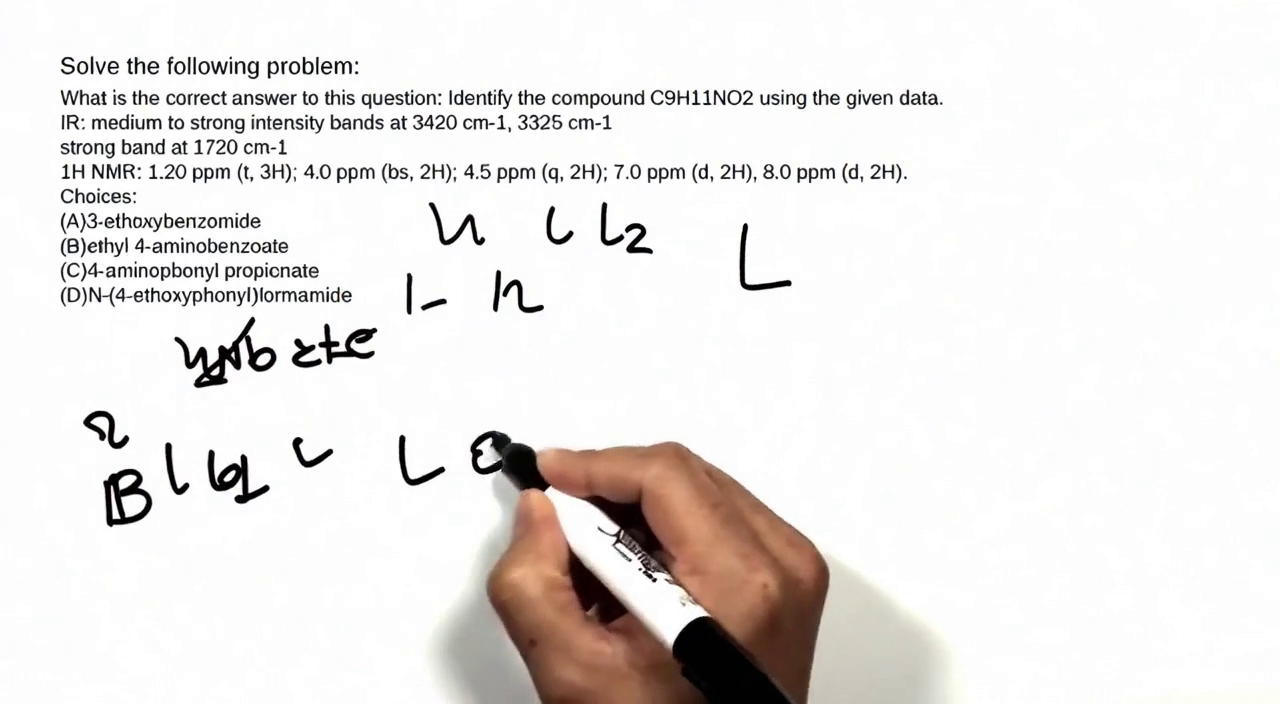}}
            \caption*{\makecell{Missing Solution Process \\ (6.96\%)}}
        \end{subfigure}
        \hfill
        \caption{Categories of Sora-2's processes in text-centric tasks.}
        \label{fig:solution_process}
    \end{minipage}%
    \hfill 
    \begin{minipage}[t]{0.44\textwidth}
        \centering
        \captionof{table}{Wan 2.5's performance on text-centric tasks with and without prompt rewriting. Its reasoning ability almost vanishes when the prompt rewriter model is disabled, indicating that the rewriter solves the reasoning problems for the video generation \mbox{component.}}
        \vspace{5pt}
        {
        \renewcommand{\arraystretch}{1.2}
        \begin{tabular}{lccc}
            \toprule
            \textbf{Dataset} & \makecell{\textbf{Prompt} \\ \textbf{Rewrite}} & \textbf{Last Frame} & \textbf{Audio} \\
            \midrule
            \multirow{2}{*}{GSM8K} & \xmark & 0.0 & 0.0 \\
             & \cmark & \textbf{78.4} & \textbf{31.9} \\
            \midrule
            \multirow{2}{*}{MMLU} & \xmark & 0.0 & 0.0 \\
             & \cmark & \textbf{74.1} & \textbf{50.00} \\
            \midrule
            \multirow{2}{*}{MMMU} & \xmark & 2.0 & 0.0 \\
             & \cmark & \textbf{47.0} & \textbf{14.0} \\
            \bottomrule
        \end{tabular}
        }
        \label{tab:wan2.5_text_reason_test}
    \end{minipage}
\end{figure}

The examples and ratios are in Figure~\ref{fig:solution_process}. We find that Sora-2 struggles to generate coherent reasoning processes in the video. Only 13.91\% of the solutions are fully correct. A large proportion of solutions (43.48\%) are unreadable or incorrect. This suggests that Sora-2 has difficulties in giving a clear and correct reasoning process via video generation.

\subsubsection{Source of Text-Centric Reasoning Ability}
\label{sec:source_text_reasoning_ability}

Sora-2's text-centric reasoning (Section~\ref{sec:text_centric_eval_results}) might come from an internal prompt rewriter. Because we cannot control Sora-2's internal rewriter, we use Wan 2.5~\citep{wan2025wan}, which provides a parameter for enabling or disabling prompt rewriting. We test Wan 2.5 on subsets of GSM8K, MMLU and MMMU under both settings.

The results (Table~\ref{tab:wan2.5_text_reason_test}) show a stark contrast: Wan 2.5 achieves nearly zero accuracy without the rewriter but improves dramatically with it, indicating its text reasoning relies almost entirely on the rewriter. An example in Appendix~\ref{app:wan_prompt_rewrite_case} shows how the rewriter turns a text problem into clear solution steps for video generation. This suggests Sora-2's text reasoning might also come from an internal prompt rewriter.

\begin{tcolorbox}[
    colback=blue!10,
    colframe=blue!50,
    boxrule=1pt,
    arc=3mm,
    left=5pt,
    right=5pt,
    top=5pt,
    bottom=5pt,breakable
]
\textbf{Takeaway 7}\quad We analyzed the source of Sora-2's text-centric reasoning capabilities. On adapted math problems, Sora-2 maintains performance comparable to the original test set. We also found that \mbox{Sora-2} struggles to generate coherent reasoning steps in the video. Finally, through comparative experiments with Wan 2.5, we speculate that Sora-2's text-centric reasoning ability originates from its \textbf{prompt rewriter}.
\end{tcolorbox}

\section{Related Work}

\textbf{Video Generation Model}: The field of video generation is advancing rapidly. Early models like OpenAI's Sora are the ``GPT-1 moment~\citep{radford2018improving}'' for video, and now newer versions like Sora-2 have made a huge leap forward. Sora-2 can create more realistic and controllable videos that are physically accurate and even include synchronized dialogue and sound effects.
Besides Sora, other powerful but closed-source models are pushing the industry forward. Companies like Runway, with its Gen-3 model~\citep{RunwayGen3Alpha}, Pika Labs, Luma AI, and Google DeepMind's Veo~\citep{GoogleVeo3ModelPage} series are all creating impressive, high-quality videos. However, because these models are proprietary, they are not widely available for researchers to study and build upon.
To counter this, a movement of open-source alternatives is growing. Projects like Stable Video Diffusion~\citep{blattmann2023stablevideodiffusionscaling}, Hunyan-Video~\citep{kong2025hunyuanvideosystematicframeworklarge}, and the Wan series~\citep{wan2025wan} are making video generation technology accessible to everyone. 

\textbf{Reasoning Paradigm Transfer}: Chain-of-Thought (CoT) significantly improves the reasoning ability of large language models (LLMs)~\citep{wei2022chain,yeo2025demystifyinglongchainofthoughtreasoning,wang2023selfconsistencyimproveschainthought,guo2025deepseek,zhang2025thinkingvideosmultimodaltoolaugmented}.
Large-scale reinforcement learning incentivizes LLMs to think productively using their CoT~\citep{openai2025learningtoreason,guo2025deepseek,zhao2024exploring}.
o3 and o4-mini further extend this capability by natively ``Thinking with Images'' in their CoT, which involves directly cropping, zooming, and rotating images~\citep{openai2025o3systemcard}. 
``Thinking with Images~\citep{openai2024gpt4o,openai2025o3systemcard,li2024enhancingadvancedvisualreasoning,zhang2024multimodalchainofthoughtreasoninglanguage}'' is a paradigm that outputs images in CoT to help VLMs reason better, largely improving the VLMs' reasoning abilities~\citep{tong2025game}. Recently, unified multimodal understanding and generating models have appeared~\citep{deng2025emerging,cui2025emu3,zhang2025vitcotvideotextinterleavedchainofthought,zhang2025rewatchr1boostingcomplexvideo,xin2025lumina}. They potentially achieve ``Thinking with Images'' through text and image interleaved reasoning. 

\textbf{Evaluation of Video Generation Reasoning}: Video-generation-based reasoning has only recently begun to be explored~\citep{wiedemer2025video,guo2025video}.
\citeauthor{wiedemer2025video} show that Veo 3 can solve many tasks it was not specifically trained for. These abilities span perceiving, modeling, and manipulating the visual world, enabling early forms of video-based reasoning. Their evaluations include tasks such as maze solving and visual symmetry.

However, existing works~\citep{wiedemer2025video,guo2025video} differ from our focus in several key aspects:
(1) Vision-centric scope: Their evaluations primarily focus on vision-centric reasoning tasks and do not extend to text-centric or broader multimodal reasoning settings.
(2) Case-based evaluation setup: These works include both qualitative demonstrations and several quantitative evaluations. However, the evaluations are conducted on a limited number of manually curated scenes or canonical examples with restricted diversity. As a result, each task is tested on relatively small sample sizes, making it challenging to assess generalization or statistical robustness.
(3) Lack of systematic comparison with VLMs: These works do not provide a systematic comparison with SOTA Vision-Language Models (VLMs) across diverse task categories, leaving the relative strengths of video vs. vision-language models underexplored.

Our work complements these directions with the following contributions:
(1) Unified multimodal reasoning paradigm: We evaluate video models not only on vision-centric tasks but also on text-centric and multimodal reasoning tasks, demonstrating that video generation may serve as a general multimodal reasoning paradigm rather than a purely visual one.
(2) Systematic and verifiable benchmark construction: We systematically construct the VideoThinkBench where large numbers of test cases can be generated in batches using a program. Most of the vision-centric tasks we have designed are verifiable.
(3) Systematic comparison with VLMs: We conduct comprehensive comparisons with SOTA VLMs, providing the first systematic study of how ``Thinking with Video'' behaves relative to ``Thinking with Images''.
In summary, we propose ``Thinking with Video'' as a new paradigm with the potential to unify multimodal reasoning. Furthermore, we find that video model reasoning can be enhanced through few-shot learning and test time scaling (self-consistency).

\section{Conclusion}

In this paper, we introduce a reasoning paradigm termed ``Thinking with Video.'' We evaluate Sora-2 on our newly constructed VideoThinkBench. Our analysis shows that Sora-2 is inherently suitable for human-like reasoning through drawing and imagination. Furthermore, it demonstrates the potential to perform textual reasoning through video frames, enabling more unified multimodal understanding and generation. Thus, ``Thinking with Video'' is potentially a unified multimodal reasoning paradigm.

\section{Limitations and Future Work}

We primarily evaluate Sora-2's reasoning abilities among video generation models. Sora-2 is not open-source, limiting the analysis of its internal mechanisms.

For future evaluation work, we plan to include more video generation models, especially open-source models. This allows for a deeper analysis of their internal mechanisms. Meanwhile, there are other capabilities of video models worth exploring.

To enhance the reasoning abilities of video models through training, a promising direction is to scale up the verifiable tasks in VideoThinkBench via Reinforcement Learning with Verifiable Rewards (RLVR), thereby enhancing models' ``Thinking with Video'' capabilities.

Regarding unified multimodal training for video models, we will explore converting textual corpora into video-form training data (e.g., by generating the next word frame-by-frame to simulate whiteboard handwriting). The idea is that by pretraining video generation models on such text-generation tasks, they can acquire textual world knowledge. Ultimately, with large-scale image-text data training, these models might achieve unified multimodal understanding and generation.

\section*{Acknowledgement}

This work is in part supported by the New Generation Artificial Intelligence-National Science and \mbox{Technology} Major Project (2025ZD0123502) and the National Natural Science Foundation of China (No. 62521004).


\bibliographystyle{plainnat}
\bibliography{main}

\clearpage


\beginappendix

\startcontents[app]
\begingroup
  \printcontents[app]{}{1}{}
\endgroup



\section{More about VideoThinkBench}

\subsection{Detailed Sample Distribution}
\label{app:bench_sample_distribution}

\begin{figure*}[t]
    \centering
    \begin{subfigure}[t]{0.49\textwidth}
        \centering
        \includegraphics[width=\linewidth]{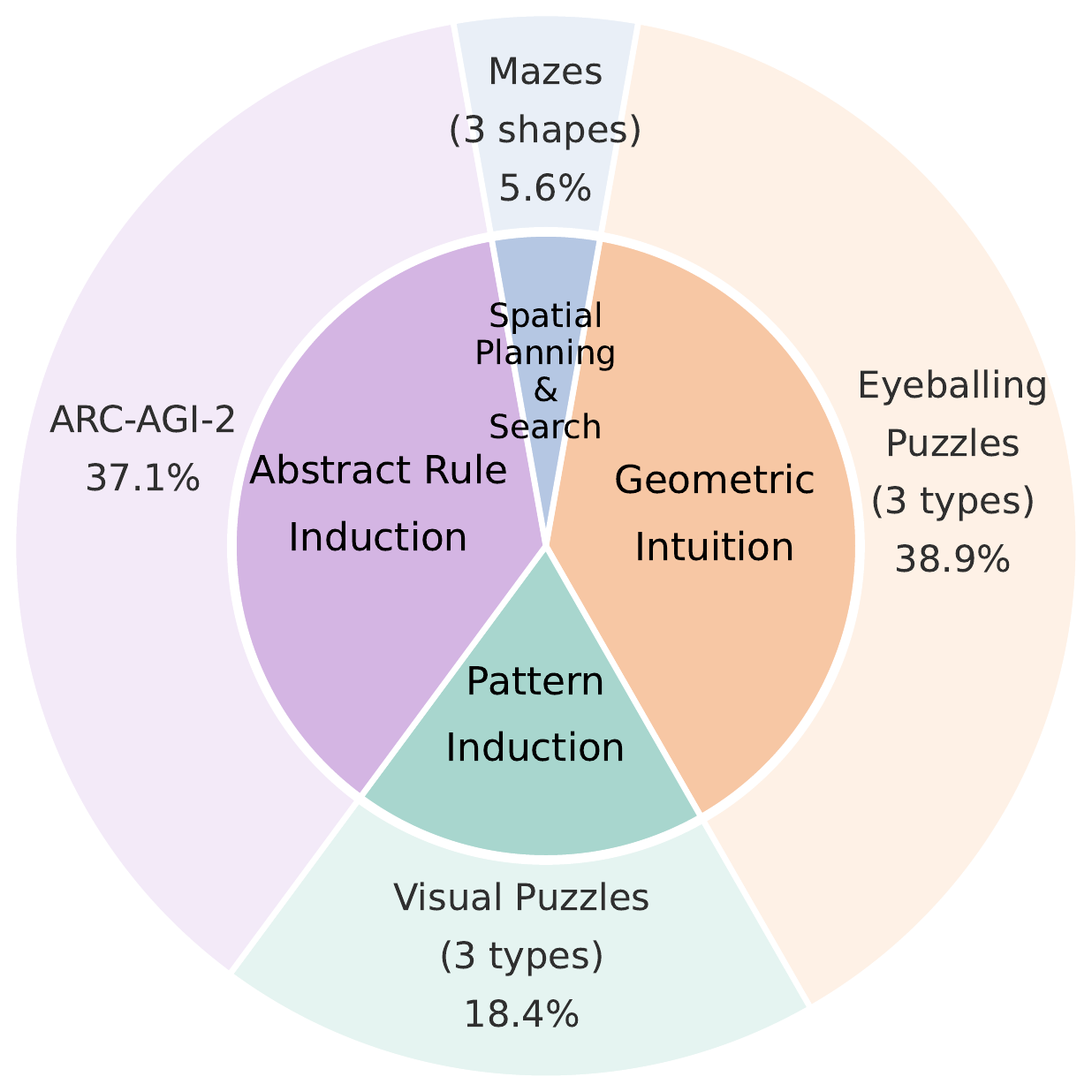}
        \caption{\textbf{Vision-centric reasoning tasks}}
        \label{fig:vision-centric_pie}
    \end{subfigure}
    \hfill
    \begin{subfigure}[t]{0.49\textwidth}
        \centering
        \includegraphics[width=\linewidth]{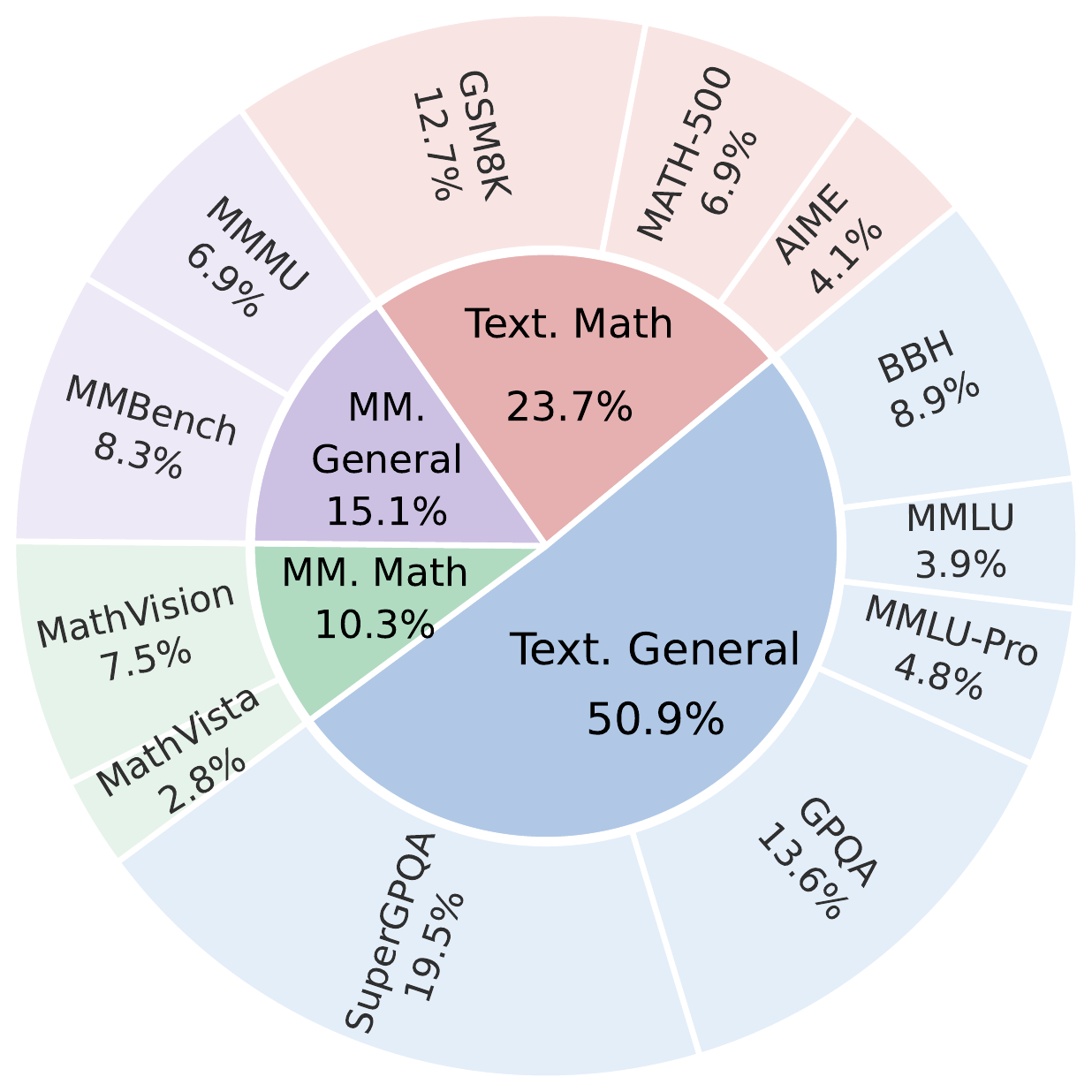}
        \caption{\textbf{Text-centric reasoning tasks}}
        \label{fig:text-centric_pie}
    \end{subfigure}
    \caption{Task composition and distribution of Video Thinking Benchmark (VideoThinkBench). (a) Vision-centric tasks contain tasks that we design (e.g., Eyeballing Puzzles) and tasks adapted from existing benchmarks (e.g., ARC-AGI-2), evaluating four abilities. (b) Text-centric tasks consist of subsets sampled from text-only and multimodal reasoning benchmarks, adapted for video generation reasoning. The former contains math reasoning (Text. Math) and general knowledge reasoning (Text. General) benchmarks and the latter also contains math reasoning (MM. Math) and general knowledge reasoning (MM. General) benchmarks.}
    \label{fig:VGR-Bench_pie}
\end{figure*}

VideoThinkBench contains 4,149 test samples in total. Vision-centric tasks contain 2,696 samples and text-centric tasks contain 1,453 samples in total. For text-centric tasks, we sampled a subset from most of the selected benchmarks for evaluation cost control. Task distribution is illustrated in Figures~\ref{fig:vision-centric_pie} and~\ref{fig:text-centric_pie}, with detailed statistics listed below.

\textbf{Vision-Centric Tasks}: Eyeballing Puzzles (1,050), Visual Puzzles (496), ARC-AGI-2 (1,000), Mazes (150).

\textbf{Text-Centric Tasks}

\begin{itemize}[leftmargin=*]
    \item Text-Only Math Reasoning (345 samples): GSM8K (185)~\citep{cobbe2021gsm8k}; MATH-500 (100)~\citep{cobbe2021training}; AIME24 (30); AIME25~(30).
    \item Text-Only General Knowledge Reasoning (739 samples): BBH (130)~\citep{suzgun2022challengingbigbenchtaskschainofthought}; MMLU (57)~\citep{hendrycks2020measuring}; MMLU-Pro (70)~\citep{wang2024mmlu}; GPQA-diamond (198)~\citep{rein2024gpqa}; SuperGPQA-easy (284)~\citep{du2025supergpqa}.
    \item Multimodal Reasoning (369 samples): MathVista (40)~\citep{lu2023mathvista}; MathVision (109)~\citep{wang2024measuring}; MMBench~(120)~\citep{liu2024mmbench}; MMMU (100)~\citep{yue2024mmmu}.
\end{itemize}

\begin{figure*}[t]
    \centering 
    \includegraphics[width=0.87\textwidth]{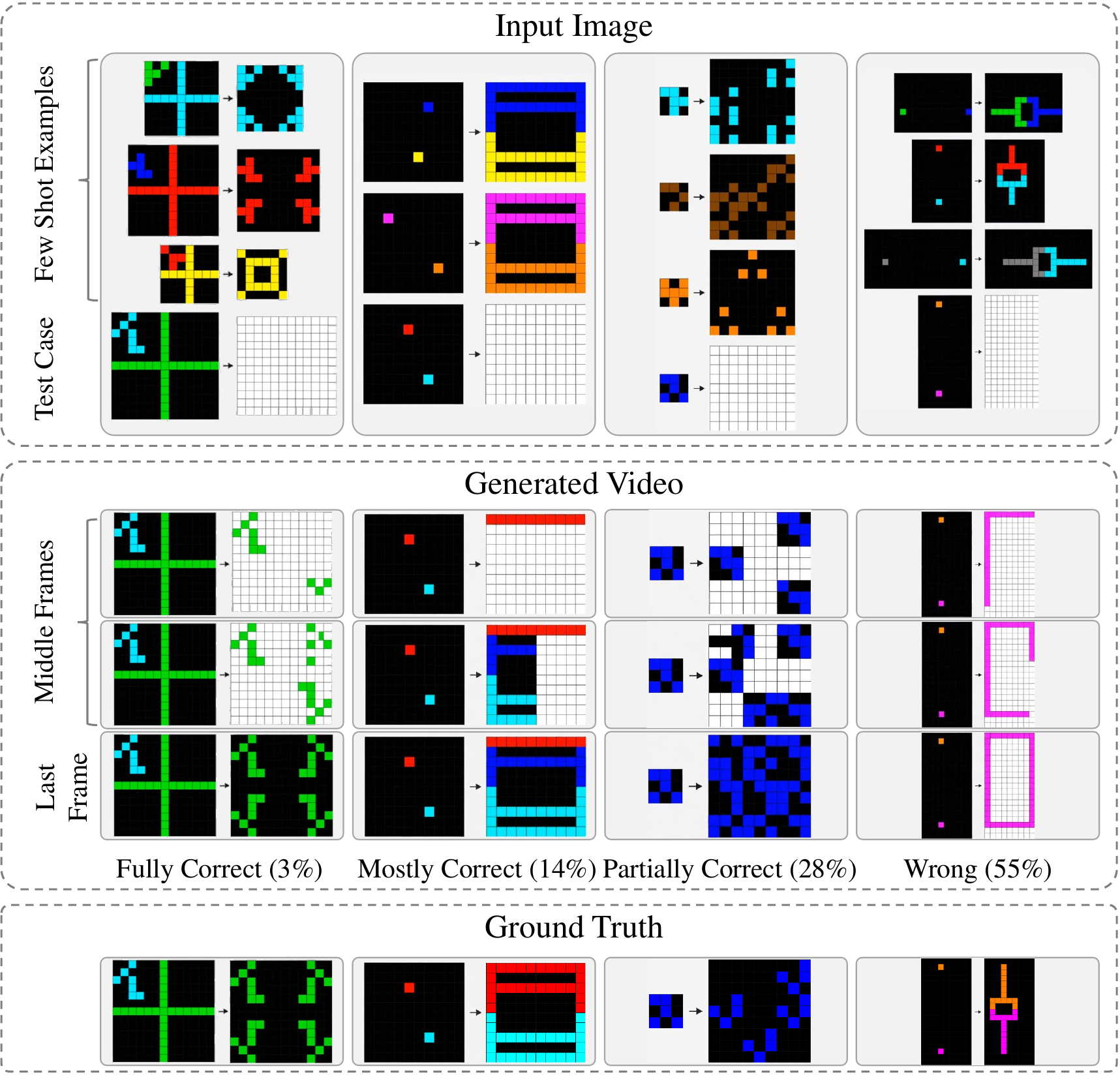}
    \caption{\textbf{Examples of Sora-2 trying to solve ARC-AGI-2.} ARC-AGI-2 is a benchmark targeting few-shot, inductive reasoning over abstract pattern transformations. Sora-2 is expected to deduce the transformation rule from the examples and use it to generate the output grid for the test case. When manually evaluating, we classify the correctness of the answer from Sora-2 into 4 categories. Details: Section~\ref{sec:arc_agi_2}.}
    \label{fig:case_arcagi2}
\end{figure*}

\subsection{Mini Test Set}
\label{app:mini_test_set}

We construct a \textbf{mini test set} to reduce the evaluation cost, making our benchmark easier for researchers to use. This mini test set is a subset of the full set and covers all the benchmark tasks, with 750 test samples in total:

\textbf{Vision-Centric Tasks} (500 samples): Eyeballing Puzzles (210, 10 per task), Visual Puzzles (100, 10 per task), ARC-AGI-2 (140), Mazes (50, covering three maze shapes); \textbf{Text-Centric Tasks}: 250 samples of the full set.

\section{More Evaluation Results}
\label{app:more_evaluation_results}

We test more models on the \textbf{mini test set} of VideoThinkBench (Appendix~\ref{app:mini_test_set}), including more video generation models (Seedance 1.0 Pro~\citep{gao2025seedance}, MiniMax Hailuo 2.3~\citep{minimax2025hailuo23}, Wan2.2-TI2V-5B~\citep{wan2025wan} and MOVA~\citep{team2026mova}), the image generation models (Nano Banana Pro~\citep{google2025gemini3proimage}, Seedream 4.5~\citep{seedream2025seedream}, GPT Image 1.5~\citep{openai2025gptimage15}, BAGEL~\citep{deng2025emerging} and Qwen-Image-Edit-2511~\citep{wu2025qwen}), and Qwen3-VL series~\citep{bai2025qwen3}. The results are shown in Table~\ref{tab:minitest_vision_centric} and Table~\ref{tab:minitest_text_centric}.

\begin{table*}[t]
    \centering
    \caption{Accuracy (\%) of different models across the vision-centric tasks of VideoThinkBench \textbf{(mini test set)}.}
    \resizebox{\linewidth}{!}{%
    \begin{tabular}{l c ccc ccc c ccc}
        \toprule
        \multirow{2}{*}{\textbf{Model}} & \multirow{2}{*}{\textbf{Average}} & \multicolumn{3}{c}{\textbf{Eyeballing Puzzles}} & \multicolumn{3}{c}{\textbf{Visual Puzzles}} & \multirow{2}{*}{\textbf{ARC-AGI-2}} & \multicolumn{3}{c}{\textbf{Mazes}} \\
        \cmidrule(lr){3-5} \cmidrule(lr){6-8} \cmidrule(lr){10-12}
        & & Point & Line & Shape & Symmetry & Gradient & Comp. & & Square & Hexagon & Circle \\
        \midrule
        \multicolumn{12}{c}{\textbf{Video Generation Models}} \\
        \midrule
        Sora-2        & \textbf{31.6} & 50.0 & 35.0 & 25.0 & 80.0 & 35.0 & 53.0 & 2.8 & 35.3 & 0.0 & 0.0 \\
        Veo 3.1       & \textbf{27.7} & 34.4 & 24.3 & 30.0 & 77.5 & 40.0 & 70.0 & 0.7 & 0.0  & 0.0 & 0.0 \\
        MiniMax Hailuo 2.3    & \textbf{25.9} & 36.7 & 34.3 & 27.5 & 72.5 & 45.0 & 42.5 & 0.0 & 0.0  & 0.0 & 0.0 \\
        MOVA-360p     & \textbf{13.4} & 23.3 & 25.7 & 25.0 & 45.0 & 0.0  & 15.0 & 0.0 & 0.0  & 0.0 & 0.0 \\
        Seedance 1.0 Pro    & \textbf{12.4} & 22.2 & 24.3 & 35.0 & 25.0 & 10.0 & 7.5  & 0.0 & 0.0  & 0.0 & 0.0 \\
        MOVA-720p     & \textbf{11.8} & 32.2 & 18.6 & 25.0 & 30.0 & 0.0  & 12.5 & 0.0 & 0.0  & 0.0 & 0.0 \\
        Wan2.2-TI2V-5B       & \textbf{7.3}  & 17.8 & 10.0 & 20.0 & 7.5  & 10.0 & 7.5  & 0.7 & 0.0  & 0.0 & 0.0 \\
        \midrule
        \multicolumn{12}{c}{\textbf{Image Generation Models}} \\
        \midrule
        Nano Banana Pro        & \textbf{29.8} & 24.0 & 30.0 & 35.0 & 85.0 & 50.0 & 73.0 & 0.7 & 0.0 & 0.0 & 0.0 \\
        Seedream 4.5         & \textbf{24.4} & 25.6 & 16.3 & 30.0 & 75.0 & 35.0 & 62.5 & 0.0 & 0.0 & 0.0 & 0.0 \\
        GPT Image 1.5          & \textbf{19.2} & 24.4 & 15.0 & 17.5 & 38.0 & 50.0 & 47.5 & 0.0 & 0.0 & 0.0 & 0.0 \\
        Qwen-Image-Edit-2511 & \textbf{14.9} & 30.0 & 23.8 & 27.5 & 25.0 & 35.0 & 7.5  & 0.0 & 0.0 & 0.0 & 0.0 \\
        BAGEL (Image Output) & \textbf{7.7}  & 24.4 & 12.5 & 25.0 & 5.0  & 0.0  & 10.0 & 0.0 & 0.0 & 0.0 & 0.0 \\
        \midrule
        \multicolumn{12}{c}{\textbf{Vision-Language Models}} \\
        \midrule
        Claude Sonnet 4.5    & \textbf{37.3} & 40.0 & 34.0 & 60.0 & 75.0 & 75.0 & 83.0 & 5.7 & 0.0 & 0.0 & 0.0 \\
        Gemini 2.5 Pro       & \textbf{35.6} & 33.0 & 23.0 & 40.0 & 95.0 & 95.0 & 68.0 & 2.1 & 0.0 & 0.0 & 0.0 \\
        GPT-5 high            & \textbf{35.5} & 39.0 & 30.0 & 23.0 & 98.0 & 80.0 & 85.0 & 0.0 & 0.0 & 0.0 & 0.0 \\
        Qwen3-VL-235B-A22B   & \textbf{30.2} & 24.0 & 17.0 & 30.0 & 93.0 & 55.0 & 83.0 & 0.0 & 0.0 & 0.0 & 0.0 \\
        Qwen3-VL-32B         & \textbf{29.6} & 33.0 & 21.0 & 20.0 & 85.0 & 55.0 & 78.0 & 4.1 & 0.0 & 0.0 & 0.0 \\
        Qwen3-VL-Plus        & \textbf{29.4} & 32.0 & 29.0 & 30.0 & 90.0 & 35.0 & 78.0 & 0.0 & 0.0 & 0.0 & 0.0 \\
        \bottomrule
    \end{tabular}
    }
    \label{tab:minitest_vision_centric}
\end{table*}

\begin{table*}[!t]
    \centering
    \caption{Accuracy (\%) of different models across the text-centric tasks of VideoThinkBench \textbf{(mini test set)}.}
    \resizebox{\linewidth}{!}{%
    \begin{tabular}{l c cccc ccccc cc cc}
        \toprule
        \multirow{2}{*}{\textbf{Model}} & \multirow{2}{*}{\textbf{Average}} & \multicolumn{4}{c}{\textbf{Text-Only Math Reasoning}} & \multicolumn{5}{c}{\textbf{Text-Only General Reasoning}} & \multicolumn{2}{c}{\textbf{MM. Math Reason.}} & \multicolumn{2}{c}{\textbf{MM. General Reason.}} \\
        \cmidrule(lr){3-6} \cmidrule(lr){7-11} \cmidrule(lr){12-13} \cmidrule(lr){14-15}
        & & GSM8K & MATH-500 & AIME24 & AIME25 & BBH & MMLU & MMLU-Pro & GPQA & SuperGPQA & MathVista & MathVision & MMBench & MMMU \\
        \midrule
        \multicolumn{15}{c}{\textbf{Video Generation Models}} \\
        \midrule
        Sora-2 (Audio)                 & \textbf{67.6} & 100.0 & 90.0 & 50.0 & 40.0 & 76.9 & 66.7 & 73.3 & 56.0 & 45.7 & 75.0 & 45.0 & 90.0 & 70.0 \\
        Sora-2 (Last Frame)            & \textbf{57.1} & 76.7  & 65.0 & 40.0 & 30.0 & 69.2 & 66.7 & 73.3 & 52.0 & 54.3 & 70.0 & 45.0 & 60.0 & 40.0 \\
        Veo 3.1 (Last Frame)           & \textbf{48.3} & 80.0 & 70.0 & 50.0 & 20.0 & 61.5 & 16.7 & 60.0 & 52.0 & 42.9 & 50.0 & 45.0 & 35.0 & 45.0 \\
        Veo 3.1 (Audio)                & \textbf{44.5} & 93.3 & 80.0 & 50.0 & 20.0 & 61.5 & 41.7 & 80.0 & 40.0 & 51.4 & 25.0 & 5.0  & 20.0 & 10.0 \\
        MiniMax Hailuo 2.3            & \textbf{38.4} & 76.6 & 40.0 & 10.0 & 20.0 & 61.5 & 33.3 & 86.6 & 16.0 & 65.7 & 30.0 & 10.0 & 30.0 & 20.0 \\
        MOVA-720p (Last Frame)         & \textbf{12.5} & 30.0 & 35.0 & 20.0 & 0.0  & 15.4 & 0.0  & 6.7  & 8.0  & 2.9  & 0.0  & 25.0 & 10.0 & 10.0 \\
        MOVA-360p (Last Frame)         & \textbf{10.4} & 20.0 & 10.0 & 0.0  & 20.0 & 15.4 & 0.0  & 0.0  & 20.0 & 0.0  & 5.0  & 30.0 & 5.0  & 10.0 \\
        MOVA-720p (Audio)              & \textbf{7.2}  & 0.0  & 0.0  & 0.0  & 0.0  & 30.8 & 16.7 & 0.0  & 8.0  & 2.9  & 0.0  & 10.0 & 20.0 & 5.0  \\
        MOVA-360p (Audio)              & \textbf{5.8}  & 0.0  & 0.0  & 0.0  & 0.0  & 30.8 & 33.3 & 0.0  & 0.0  & 5.7  & 0.0  & 0.0  & 5.0  & 0.0  \\
        Seedance 1.0 Pro        & \textbf{0.0}  & 0.0  & 0.0  & 0.0  & 0.0  & 0.0  & 0.0  & 0.0  & 0.0  & 0.0  & 0.0  & 0.0  & 0.0  & 0.0  \\
        Wan2.2-TI2V-5B                 & \textbf{0.0}  & 0.0  & 0.0  & 0.0  & 0.0  & 0.0  & 0.0  & 0.0  & 0.0  & 0.0  & 0.0  & 0.0  & 0.0  & 0.0  \\
        \midrule
        \multicolumn{15}{c}{\textbf{Image Generation Models}} \\
        \midrule
        Nano Banana Pro                  & \textbf{66.0} & 56.7 & 65.0 & 80.0 & 80.0 & 69.2 & 75.0 & 80.0 & 44.0 & 62.9 & 75.0 & 45.0 & 75.0 & 50.0 \\
        Seedream 4.5                   & \textbf{55.7} & 100.0& 80.0 & 20.0 & 10.0 & 69.2 & 75.0 & 60.0 & 36.0 & 48.6 & 55.0 & 60.0 & 55.0 & 55.0 \\
        GPT Image 1.5                  & \textbf{41.4} & 90.0 & 40.0 & 0.0  & 0.0  & 69.2 & 25.0 & 46.7 & 40.0 & 22.9 & 50.0 & 40.0 & 65.0 & 50.0 \\
        Qwen-Image-Edit-2511           & \textbf{10.5} & 0.0  & 0.0  & 0.0  & 0.0  & 15.4 & 16.7 & 6.7  & 4.0  & 8.6  & 15.0 & 15.0 & 30.0 & 25.0 \\
        BAGEL (Image Output)           & \textbf{8.9}  & 6.7  & 0.0  & 0.0  & 0.0  & 0.0  & 0.0  & 6.7  & 4.0  & 2.9  & 25.0 & 40.0 & 10.0 & 20.0 \\
        \midrule
        \multicolumn{15}{c}{\textbf{Vision-Language Models}} \\
        \midrule
        Gemini 2.5 Pro                 & \textbf{89.0} & 100.0 & 100.0 & 100.0 & 90.0 & 100.0 & 83.3 & 93.3 & 80.0 & 80.0 & 85.0 & 65.0 & 95.0 & 85.0 \\
        GPT-5 high                      & \textbf{86.6} & 100.0 & 100.0 & 100.0 & 100.0 & 100.0 & 83.3 & 93.3 & 80.0 & 83.9 & 75.0 & 55.0 & 85.0 & 70.0 \\
        Qwen3-VL-235B-A22B             & \textbf{77.6} & 100.0& 100.0& 80.0 & 50.0 & 84.6 & 58.3 & 100.0& 56.0 & 80.0 & 70.0 & 65.0 & 90.0 & 75.0 \\
        Claude Sonnet 4.5              & \textbf{77.2} & 100.0 & 100.0 & 60.0 & 40.0 & 100.0 & 83.3 & 100.0 & 60.0 & 80.0 & 80.0 & 45.0 & 80.0 & 75.0 \\
        Qwen3-VL-Plus                  & \textbf{75.8} & 100.0& 95.0 & 100.0& 70.0 & 76.9 & 66.7 & 80.0 & 64.0 & 57.1 & 65.0 & 65.0 & 80.0 & 65.0 \\
        Qwen3-VL-32B                   & \textbf{72.5} & 100.0& 95.0 & 80.0 & 50.0 & 76.9 & 66.7 & 93.3 & 40.0 & 65.7 & 75.0 & 45.0 & 90.0 & 65.0 \\
        \bottomrule
    \end{tabular}
    }
    \label{tab:minitest_text_centric}
\end{table*}

\section{Detailed Evaluation Protocols}
\label{app:evaluation_details}

This section provides comprehensive details on the evaluation protocols for all tasks in VideoThinkBench. We present the dataset construction methods, evaluation procedures, and some prompts used for both video generation models and VLM baselines.


\subsection{Generation Parameters}

For Sora-2, the video duration is 10 seconds in all the experiments. For evaluation of Wan 2.5 detailed in Section~\ref{sec:source_text_reasoning_ability}, we use the model of wan2.5-i2v-preview, setting the resolution to 480P and the duration to five seconds.

\subsection{ARC-AGI-2}
\label{appendix-sec:ARC-AGI-2}

We present four cases and the corresponding manual evaluation category in Figure~\ref{fig:case_arcagi2}.


\subsection{Mazes}
\label{sec:maze}




\begin{minipage}[t]{0.6\textwidth}
\paragraph{Dataset Construction}
We use programs to automatically construct a dataset of 150 mazes, divided equally into three distinct geometric types: square mazes, hexagon mazes, and circle mazes. For each type, we generated 50 unique instances, each with a start and end point marked by red dots. The task requires the model to generate a path from start to end while not overlapping black walls.

\vspace{0.5em} 
\paragraph{Evaluation Setup}
Evaluation is conducted automatically on the final frame of the generated video. A solution is considered successful only if it satisfies two conditions: 1) Red pixels form a continuous line connecting the start and end points. 2) No red pixel overlaps any black pixel in the input image which represents the maze walls. An attempt is marked correct only if both criteria are fully met.

\vspace{0.5em} 
\paragraph{Evaluation Results}
Sora-2's performance on the maze-solving task varied significantly depending on the maze's geometric structure. As shown in Figure~\ref{fig:maze}, it demonstrated a moderate ability to solve traditional square mazes, successfully finding a valid path in 20 out of 50 instances for a \textbf{40\% success rate}. However, the model's spatial reasoning did not extend to other geometries. For both the hexagon and circle mazes, Sora-2 failed to produce a single correct solution, resulting in a \textbf{0\% success rate} for both categories. This stark performance gap suggests that while Sora-2 can handle basic pathfinding on grid-like structures, its reasoning struggles to adapt to more complex shapes.
\end{minipage}%
\hfill
\begin{minipage}[t]{0.37\textwidth}
    \vspace{0pt} 
    \centering
    \small
    \includegraphics[width=\linewidth]{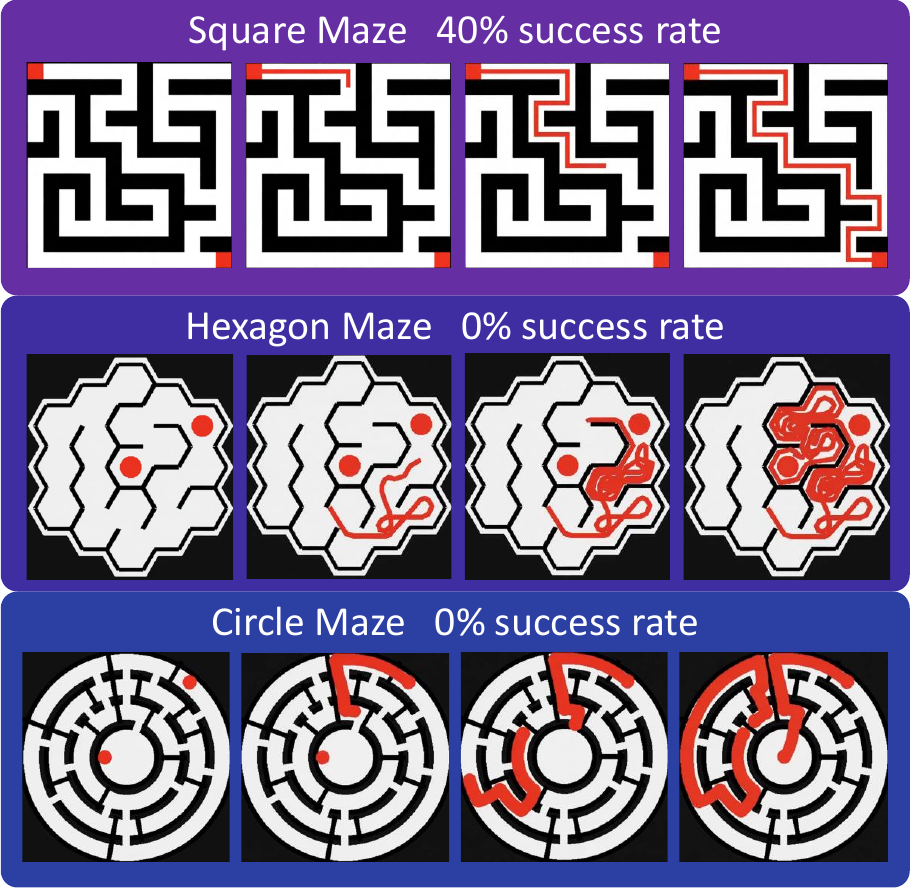} 
    \captionof{figure}{\textbf{Examples and evaluation results of Mazes.} In each quartet, the first image is the input image, and other three images are from videos generated by Sora-2. We generate and evaluate 50 samples for each type of maze. Prompt: \textit{``Draw a red path connecting two red dots without touching the black walls. In portrait. Static camera.''} Sora-2 successfully solves the square maze but fails at other two mazes. Details: Section~\ref{sec:maze}.}
    \label{fig:maze}
\end{minipage}


\vspace{2em} 




\subsection{Eyeballing Puzzles}

\begin{figure*}[t]
\centering
\includegraphics[scale=0.49]{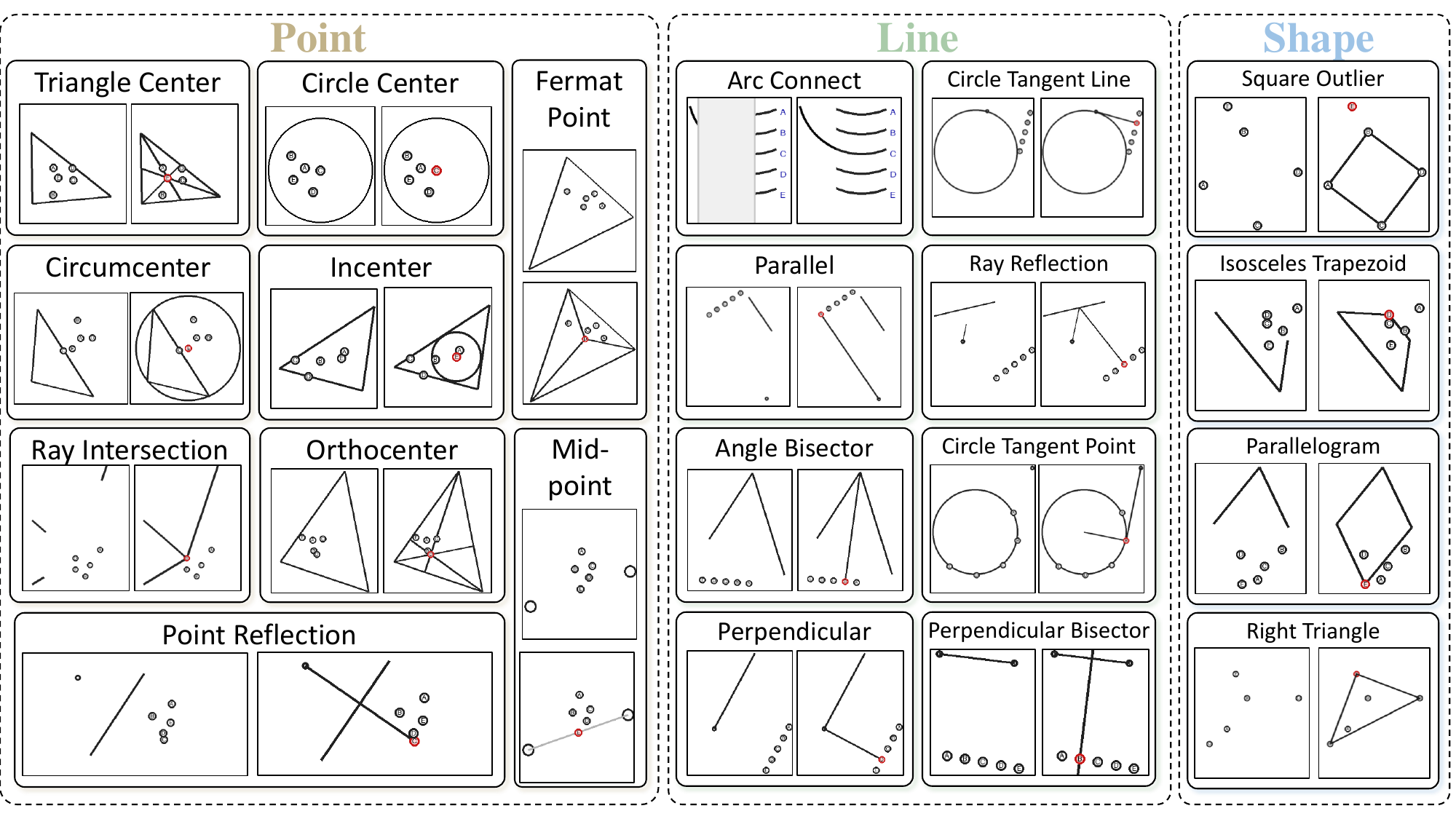}

\caption{\textbf{Overview of 21 eyeballing puzzle types.} Based on the task requirement (constructing a point, a line, or a shape), we divide the puzzle types into Point, Line, and Shape categories. For each puzzle type, an input image and corresponding ground truth image is shown. All prompts: Section~\ref{appendix_sec:eyeballing_prompts}.}
\label{fig:eyeballing_overview}
\end{figure*}

\paragraph{Overview}
\label{appendix_sec:eyeballing_overview}
We show examples of all 21 eyeballing puzzle types in Figure~\ref{fig:eyeballing_overview}.

\label{appendix_sec:eyeballing_prompts}

\paragraph{Evaluation Details}
For Sora-2, the three evaluation methods on Eyeballing Puzzles are introduced as follows:
\begin{itemize}
    \item \textbf{Audio Evaluation:} The prompt instructs model to speak out the option in phonetic alphabet (``Alpha'', ``Bravo'', ``Charlie'', ``Delta'' and ``Echo''). Audio is extracted from generated video and transcribed using whisper-1 model. Then a program finds first appearing phonetic alphabet word as the audio option. Finally, Compare the audio option with ground truth.
    
    \item \textbf{Last Frame Evaluation:} The prompt instructs model to draw a red dot on correct option. The last frame of generated video is fed to an image evaluator program that calculates average coordination of red pixels. The last frame option is the option nearest to average coordination of red pixels, or none if there are no red pixels found. Finally, Compare the last frame option with ground truth.
    
    \item \textbf{Major Frame Evaluation:} For every 5 frames in the video, one frame is extracted and fed to the image evaluator, getting option of this frame. Major frame option is the majority vote result of all chosen frames. ``None'' option is excluded from voting. Finally, Compare the Major frame option with ground truth.
\end{itemize}

\paragraph{Results in Table}
We present the results of eyeballing tasks in Table~\ref{tab:spatial_reasoning_results_table}.

\newcolumntype{C}{>{\centering\arraybackslash}X}
\begin{table*}[t]
    \centering
    \caption{Accuracy (\%) of Sora-2 using 3 evaluation methods and 3 VLMs on eyeballing tasks. The highest score in each row is highlighted in bold. Details: Section~\ref{sec:spatial_reasoning}.}
    \label{tab:spatial_reasoning_results_table}
    \small
    \setlength{\tabcolsep}{2pt} 
    \begin{tabularx}{\linewidth}{lCCCCCC}
    \toprule
    \multirow{2}{*}{\textbf{Task}} & \textbf{Sora-2} & \textbf{Sora-2} & \textbf{Sora-2} & \textbf{Gemini} & \textbf{GPT-5} & \textbf{Claude} \\
    &\textbf{Audio}&\textbf{Last Frame}&\textbf{Major Frame}&\textbf{2.5 Pro}&\textbf{high}&\textbf{Sonnet 4.5}\\
    \midrule
    \multicolumn{7}{c}{\textit{Point Tasks}} \\
    Ray Intersection       & 22.0 & 70.0 & \textbf{88.0} & 22.0 & 16.0 & 22.0 \\
    Midpoint               & 22.0 & 48.0 & 64.0 & 28.0 & 34.0 & \textbf{66.0} \\
    Circle Center          & 58.0 & 56.0 & \textbf{70.0} & 44.0 & 62.0 & 50.0 \\
    Point Reflection       & 18.0 & 22.0 & 22.0 & \textbf{30.0} & 28.0 & \textbf{30.0} \\
    Triangle Center        & 34.0 & 42.0 & \textbf{44.0} & 38.0 & 40.0 & 36.0 \\
    Incenter               & \textbf{48.0} & 30.0 & 34.0 & 32.0 & 30.0 & 34.0 \\
    Circumcenter           & 14.0 & 20.0 & 24.0 & 12.0 & \textbf{32.0} & 26.0 \\
    Orthocenter            & \textbf{32.0} & 18.0 & 26.0 & 14.0 & \textbf{32.0} & 28.0 \\
    Fermat Point           & 24.0 & 24.0 & 30.0 & 30.0 & 28.0 & \textbf{34.0} \\
    \textbf{Average}       & 30.2 & 36.7 & \textbf{44.7} & 27.8 & 33.6 & 36.2 \\
    \midrule
    \multicolumn{7}{c}{\textit{Line Tasks}} \\
    Perpendicular          & 20.0 & 38.0 & \textbf{46.0} & 8.0  & 26.0 & 14.0 \\
    Parallel               & 22.0 & 28.0 & 30.0 & 20.0 & \textbf{32.0} & \textbf{32.0} \\
    Angle Bisector         & 28.0 & 36.0 & \textbf{38.0} & 28.0 & 28.0 & 24.0 \\
    Arc Connect            & 12.0 & 56.0 & \textbf{68.0} & 20.0 & 20.0 & 12.0 \\
    Perpendicular Bisector & 22.0 & 20.0 & 40.0 & 16.0 & 30.0 & \textbf{58.0} \\
    Circle Tangent Line    & 22.0 & 20.0 & \textbf{26.0} & 22.0 & 20.0 & 22.0 \\
    Circle Tangent Point   & 18.0 & 16.0 & \textbf{24.0} & 22.0 & 18.0 & 22.0 \\
    Ray Reflection         & 28.0 & 30.0 & \textbf{32.0} & \textbf{32.0} & 18.0 & 26.0 \\
    \textbf{Average}       & 21.5 & 30.5 & \textbf{38.0} & 21.0 & 24.0 & 26.3 \\
    \midrule
    \multicolumn{7}{c}{\textit{Shape Tasks}} \\
    Square Outlier         & 54.0 & 44.0 & 54.0 & 52.0 & 54.0 & \textbf{86.0} \\
    Parallelogram          & 24.0 & 28.0 & 32.0 & 24.0 & 30.0 & \textbf{36.0} \\
    Right Triangle         & 30.0 & 14.0 & 16.0 & 38.0 & 20.0 & \textbf{60.0} \\
    Isosceles Trapezoid    & 36.0 & \textbf{42.0} & 36.0 & 24.0 & 26.0 & 20.0 \\
    \textbf{Average}       & 36.0 & 32.0 & 34.5 & 34.5 & 32.5 & \textbf{50.5} \\
    \midrule
    \textbf{Overall Average} & 28.0 & 33.4 & \textbf{40.2} & 26.5 & 29.7 & 35.1 \\
    \bottomrule
    \end{tabularx}
\end{table*}

\begin{figure*}[!t]
\centering 
\includegraphics[scale=0.46]{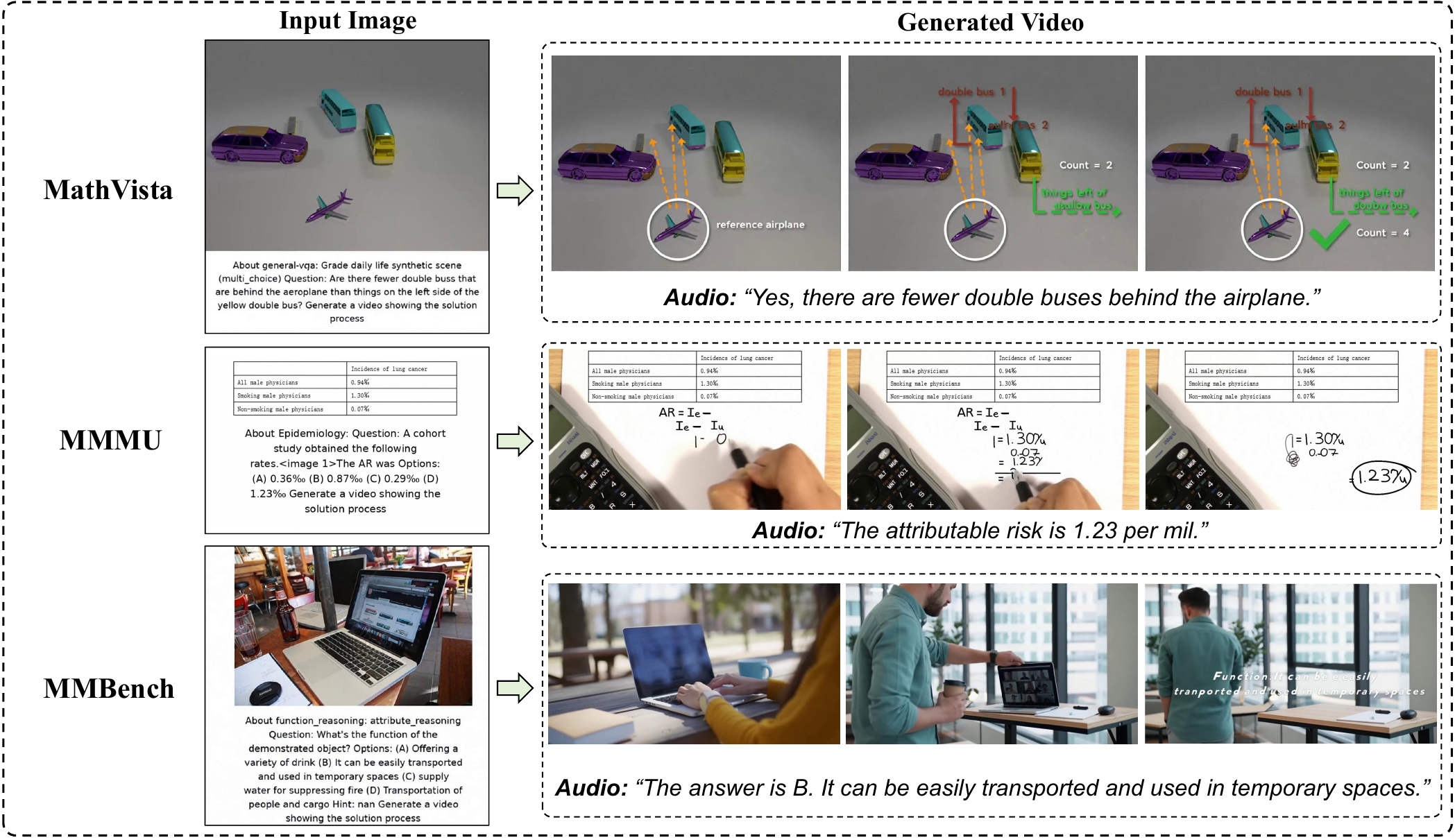}
\caption{Sora-2 solves the multimodal reasoning questions of the text-centric tasks. The input image contains the image of the original multimodal reasoning problem and the question. Similar to Figure~\ref{fig:text-centric_eval_pipeline}, a text prompt containing the question text is also input to the model.}
\label{fig:multimodal}
\end{figure*}

\subsection{Visual Puzzles}
\begin{figure*}[t]
\centering 
\includegraphics[scale=0.5]{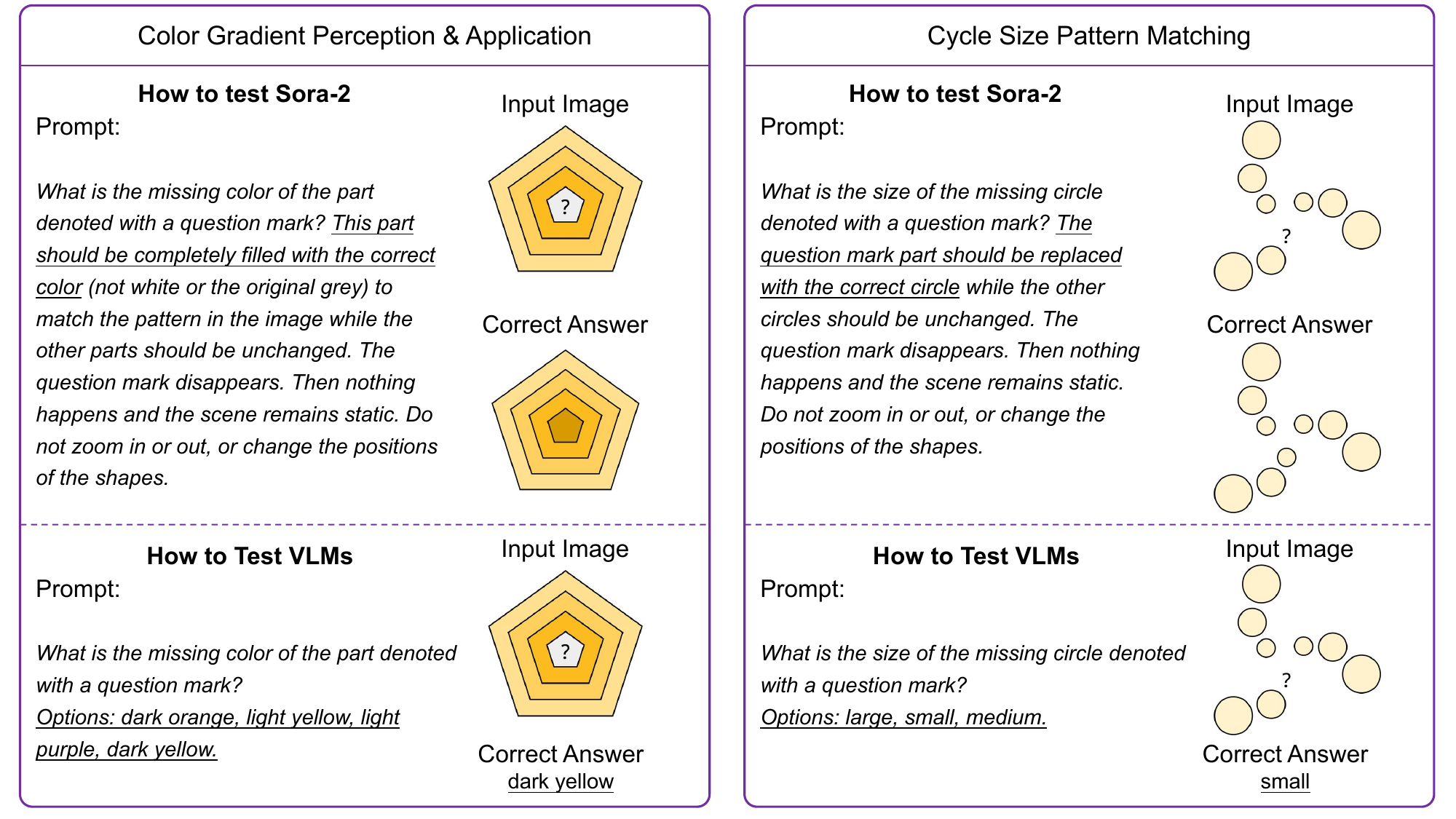}
\caption{Two examples from the five visual puzzle tasks (listed in Appendix~\ref{app:visual_puzzle_eval}) where \textbf{the tested VLMs are provided with multiple-choice options}. The VLMs only need to select the correct option while Sora-2 needs to correctly solve the tasks in the generated videos.}
\label{fig:visual_puzzle_sora_VLM_test_form_cmp}
\end{figure*}

\begin{figure*}[!t]
    \centering
    
    \begin{subfigure}[t]{0.31\textwidth}
        \centering
        \textbf{(1) Hexagon Color} 
        \begin{tabular}{@{}c@{\hfill}c@{}}
            \includegraphics[width=0.48\linewidth]{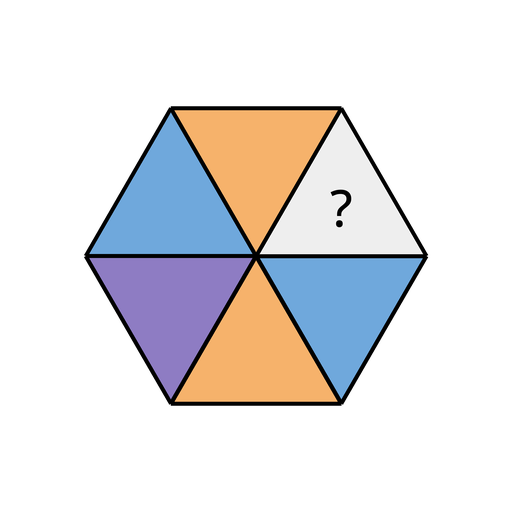} &
            \includegraphics[width=0.48\linewidth]{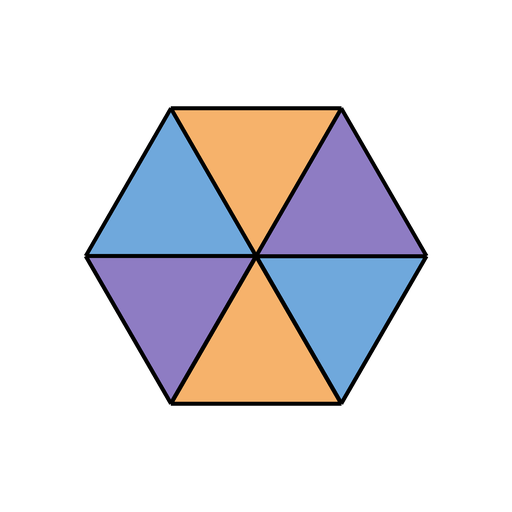} \\
            \small \makebox[0.48\linewidth][c]{Puzzle} & \small \makebox[0.48\linewidth][c]{Solution}
        \end{tabular}
    \end{subfigure}
    \hfill
    \begin{subfigure}[t]{0.31\textwidth}
        \centering
        \textbf{(2) Grid Color} 
        \begin{tabular}{@{}c@{\hfill}c@{}}
            \includegraphics[width=0.48\linewidth]{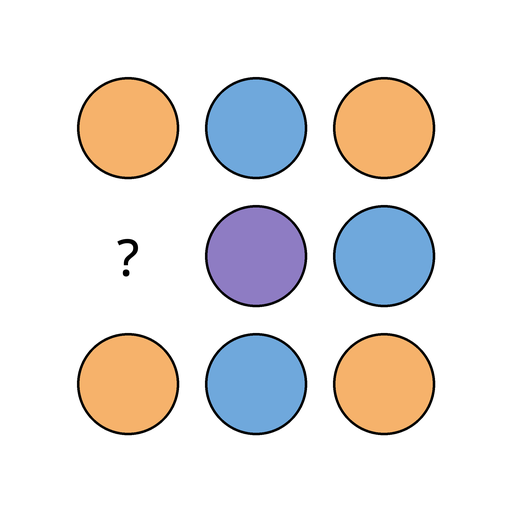} &
            \includegraphics[width=0.48\linewidth]{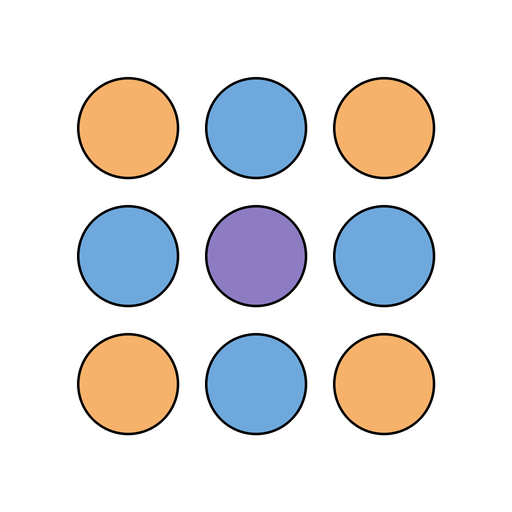} \\
            \small \makebox[0.48\linewidth][c]{Puzzle} & \small \makebox[0.48\linewidth][c]{Solution}
        \end{tabular}
    \end{subfigure}
    \hfill
    \begin{subfigure}[t]{0.31\textwidth}
        \centering
        \textbf{(3) Grid Size Pattern} 
        \begin{tabular}{@{}c@{\hfill}c@{}}
            \includegraphics[width=0.48\linewidth]{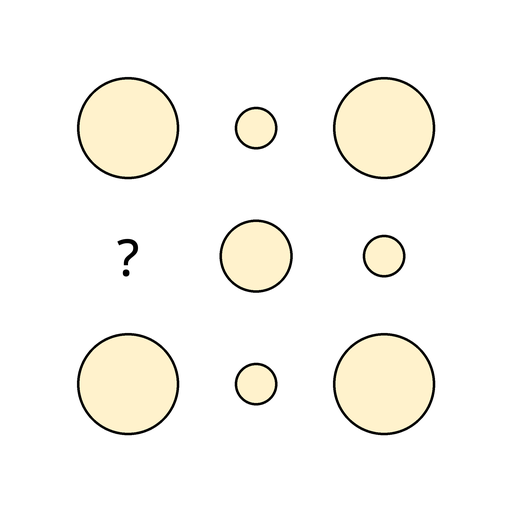} &
            \includegraphics[width=0.48\linewidth]{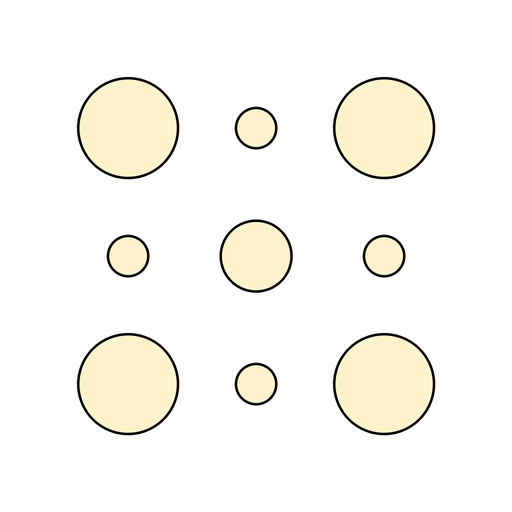} \\
            \small \makebox[0.48\linewidth][c]{Puzzle} & \small \makebox[0.48\linewidth][c]{Solution}
        \end{tabular}
    \end{subfigure}

    \vspace{3mm}

    \begin{subfigure}[t]{0.31\textwidth}
        \centering
        \textbf{(4) Reflection} 
        \begin{tabular}{@{}c@{\hfill}c@{}}
            \includegraphics[width=0.48\linewidth]{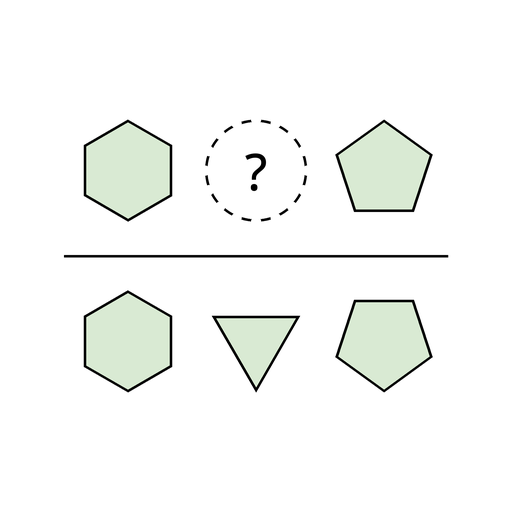} &
            \includegraphics[width=0.48\linewidth]{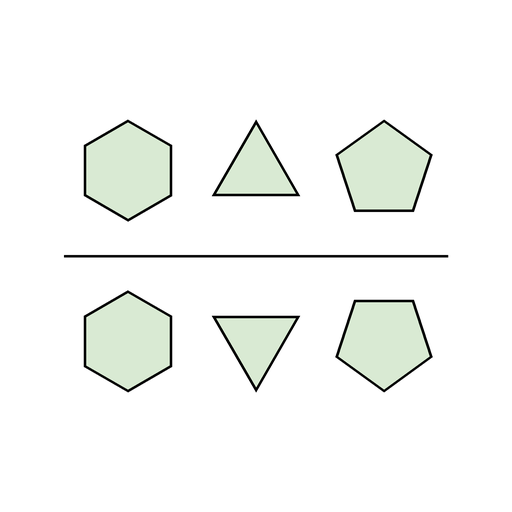} \\
            \small \makebox[0.48\linewidth][c]{Puzzle} & \small \makebox[0.48\linewidth][c]{Solution}
        \end{tabular}
    \end{subfigure}
    \hfill
    \begin{subfigure}[t]{0.31\textwidth}
        \centering
        \textbf{(5) Color Gradient} 
        \begin{tabular}{@{}c@{\hfill}c@{}}
            \includegraphics[width=0.48\linewidth]{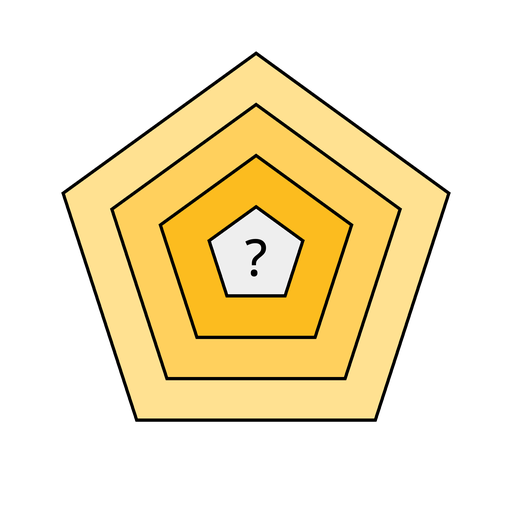} &
            \includegraphics[width=0.48\linewidth]{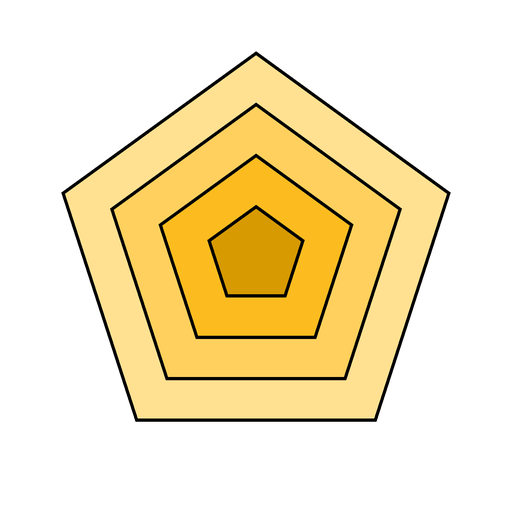} \\
            \small \makebox[0.48\linewidth][c]{Puzzle} & \small \makebox[0.48\linewidth][c]{Solution}
        \end{tabular}
    \end{subfigure}
    \hfill
    \begin{subfigure}[t]{0.31\textwidth}
        \centering
        \textbf{(6) Cycle Size Pattern} 
        \begin{tabular}{@{}c@{\hfill}c@{}}
            \includegraphics[width=0.48\linewidth]{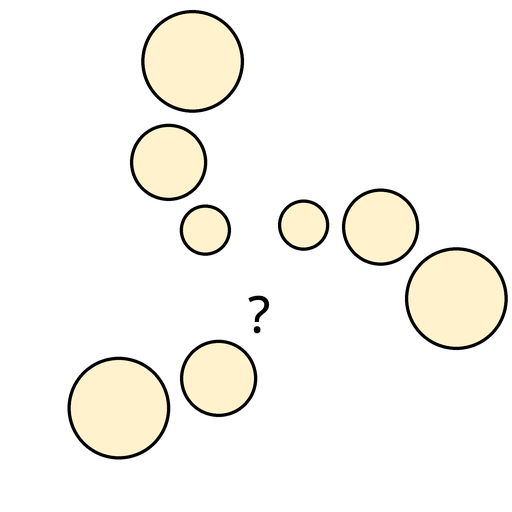} &
            \includegraphics[width=0.48\linewidth]{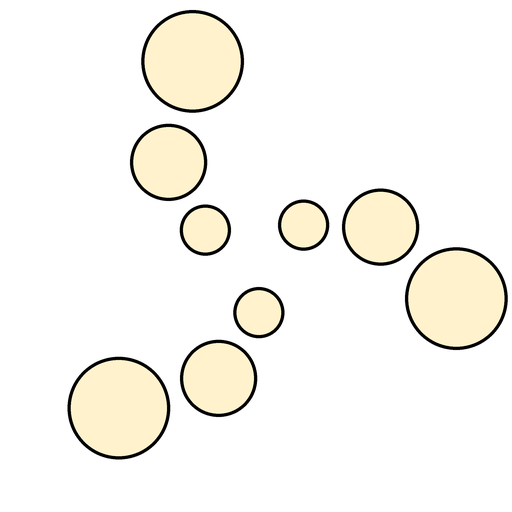} \\
            \small \makebox[0.48\linewidth][c]{Puzzle} & \small \makebox[0.48\linewidth][c]{Solution}
        \end{tabular}
    \end{subfigure}

    \vspace{3mm}

    \begin{subfigure}[t]{0.31\textwidth}
        \centering
        \textbf{(7) Shape Color} 
        \begin{tabular}{@{}c@{\hfill}c@{}}
            \includegraphics[width=0.48\linewidth]{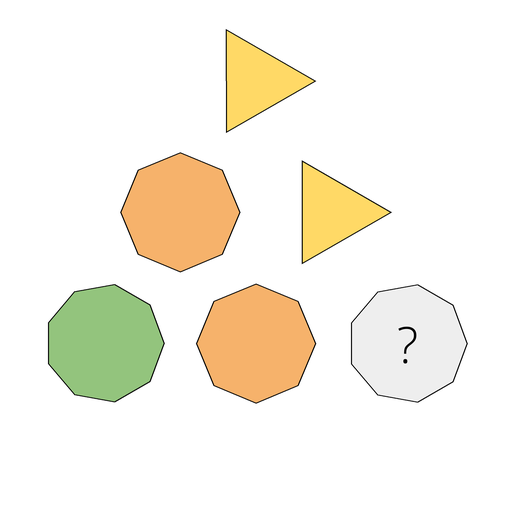} &
            \includegraphics[width=0.48\linewidth]{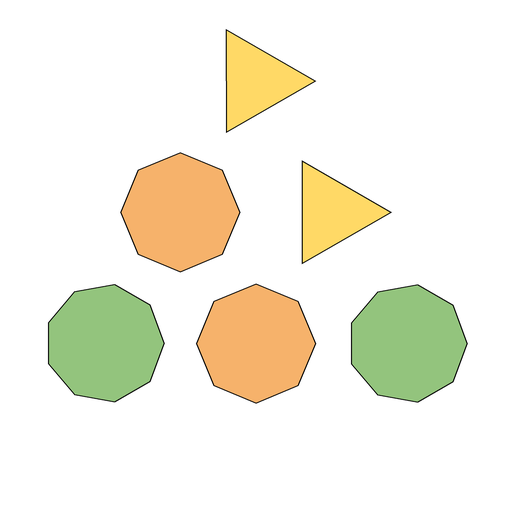} \\
            \small \makebox[0.48\linewidth][c]{Puzzle} & \small \makebox[0.48\linewidth][c]{Solution}
        \end{tabular}
    \end{subfigure}
    \hfill
    \begin{subfigure}[t]{0.31\textwidth}
        \centering
        \textbf{(8) Rectangle Height Color} 
        \begin{tabular}{@{}c@{\hfill}c@{}}
            \includegraphics[width=0.48\linewidth]{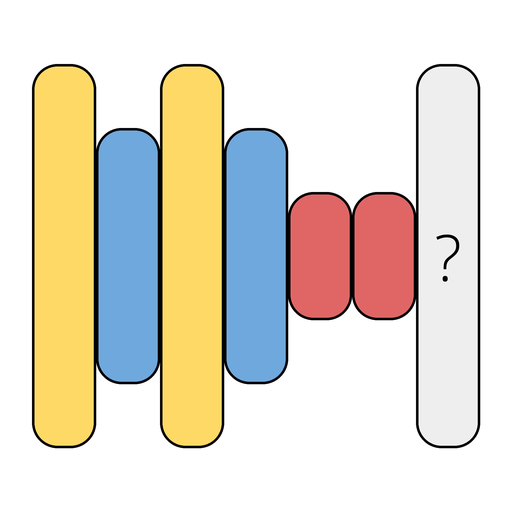} &
            \includegraphics[width=0.48\linewidth]{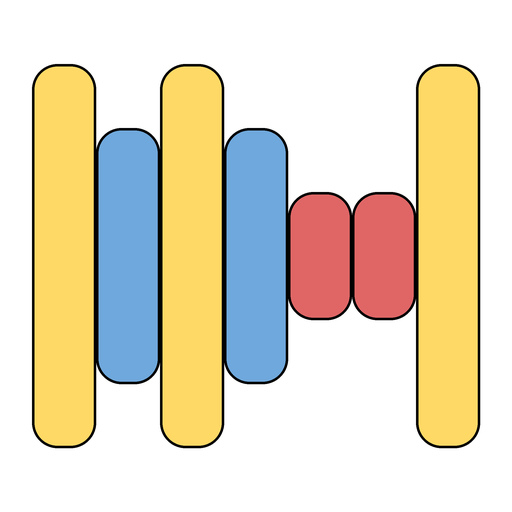} \\
            \small \makebox[0.48\linewidth][c]{Puzzle} & \small \makebox[0.48\linewidth][c]{Solution}
        \end{tabular}
    \end{subfigure}
    \hfill
    \begin{subfigure}[t]{0.31\textwidth}
        \centering
        \textbf{(9) Color Mixing} 
        \begin{tabular}{@{}c@{\hfill}c@{}}
            \includegraphics[width=0.48\linewidth]{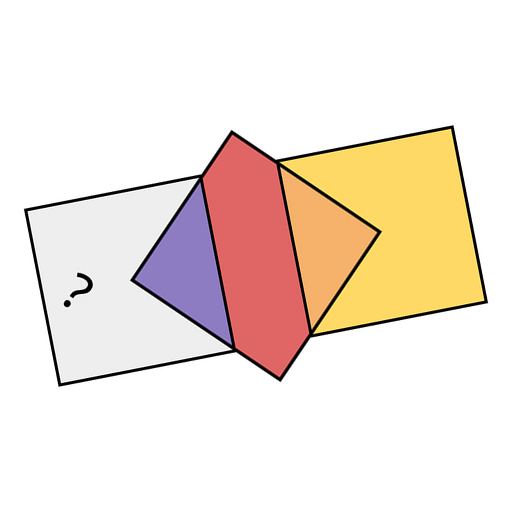} &
            \includegraphics[width=0.48\linewidth]{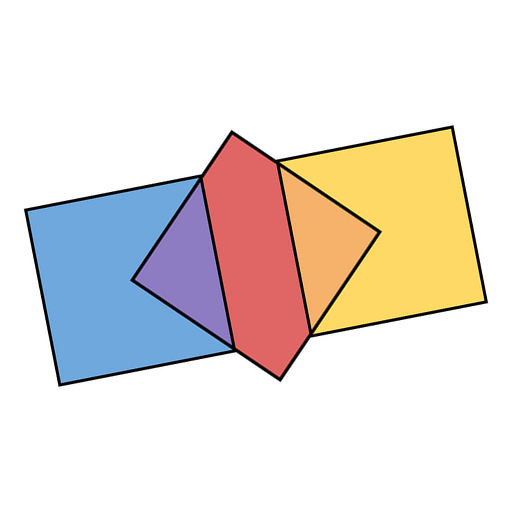} \\
            \small \makebox[0.48\linewidth][c]{Puzzle} & \small \makebox[0.48\linewidth][c]{Solution}
        \end{tabular}
    \end{subfigure}
    
    \vspace{3mm}

    \begin{subfigure}[t]{0.31\textwidth}
        \centering
        \textbf{(10) Grid Shape \& Size Pattern} 
        \begin{tabular}{@{}c@{\hfill}c@{}}
            \includegraphics[width=0.48\linewidth]{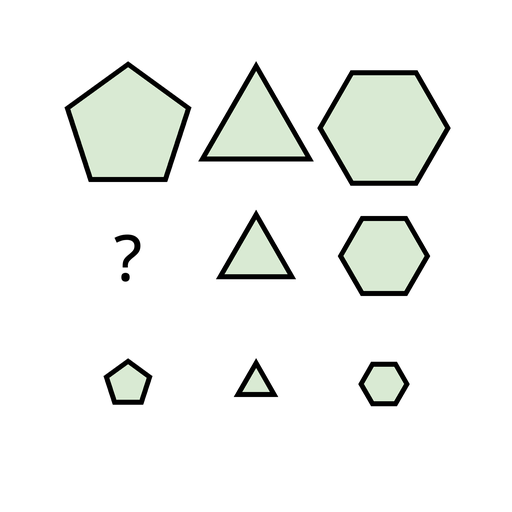} &
            \includegraphics[width=0.48\linewidth]{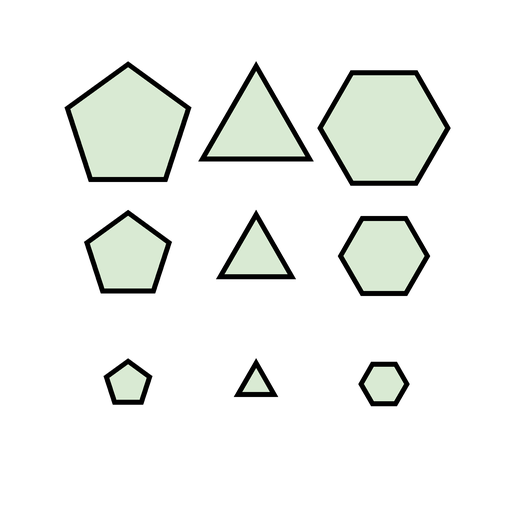} \\
            \small \makebox[0.48\linewidth][c]{Puzzle} & \small \makebox[0.48\linewidth][c]{Solution}
        \end{tabular}
    \end{subfigure}

    \caption{An overview of the 10 visual puzzle tasks evaluating inductive reasoning. The tasks are ordered by category: \textbf{Symmetry Tasks} (1)-(4), \textbf{Gradient Tasks} (5)-(6), and \textbf{Compositionality Tasks} (7)-(10). Each task displays a puzzle example and its solution.}
    \label{fig:visual_puzzles_demo}
\end{figure*}

\paragraph{Formal Definitions of the Deviation Value}
\label{app:visual_puzzle_diff}

In visual puzzles, the deviation value $Diff$ is defined to quantify the deviation between a generated video frame and the ground truth image. This metric is computed as \textbf{the sum of per-pixel differences within the puzzle area}. Formally:

\begin{equation}
    Diff = \Sigma_{(x, y) \in \text{Puzzle Area}} \delta\text{(} \text{Pixel}_{\text{gen}}\text{(}x, y\text{)}, \text{Pixel}_{\text{gt}}\text{(}x, y\text{)}\text{)}
\end{equation}

The per-pixel difference function $\delta$ is defined according to the task type:

\begin{itemize}
    \item For color-filling tasks: We calculate \textbf{the Euclidean distance in RGB space}:
        \begin{equation}
        \delta_{\text{color}}\text{(}p, q\text{)} = \sqrt{\text{(}p_r - q_r\text{)}^2\text{ + }\text{(}p_g - q_g\text{)}^2\text{ + }\text{(}p_b - q_b\text{)}^2}
        \end{equation}
        where $p$ and $q$ are the pixels from the generated and ground truth images, respectively.
    
    \item For shape-drawing tasks: The images are first converted to grayscale. Then we binarize the images and compute an \textbf{``coverage difference''}, where a pixel is considered different if its binarized color (black/white) differs:
        \begin{equation}
        \delta_{\text{shape}}\text{(}p, q\text{)} =
        \begin{cases}
        1, & \text{if } \text{Binarize(}p\text{)} \neq \text{Binarize(}q\text{)} \\
        0, & \text{otherwise}
        \end{cases}
        \end{equation}
    Here, ``$\text{Binarize}$'' uses a fixed threshold of 245. Pixels with intensity greater than this threshold are set to white (255), and others to black (0).
\end{itemize}

\paragraph{Evaluation}
\label{app:visual_puzzle_eval}

For Sora-2, we manually evaluate the performance on each of the 10 tasks, based on the selected ``best'' frames (detailed in Section~\ref{sec:visual_puzzles}). For the VLMs, we employ a rule-based evaluation by directly comparing their final answers with the ground truth answer for each test sample. For five of the 10 tasks, we provided multiple-choice options to reduce answer diversity and simplify evaluation. These five tasks are: Task 5 (Color Gradient Perception \& Application) and the four shape-drawing tasks (Tasks 3, 4, 6, 10), all of which are illustrated in Figure~\ref{fig:visual_puzzles_demo}. For the other tasks, no multiple-choice options are provided. Detailed prompts are shown in~\ref{app:visual_puzzle_prompts}.

\paragraph{Results in Table}
We present the detailed evaluation results of visual puzzles in Table~\ref{tab:visual_puzzle_result}.

\subsection{Text-Centric Tasks}



\paragraph{Human Alignment Check for Evaluation}
\label{app:human_alignment_text}

We performed a human alignment check on a sample of 173 responses across the text-centric tasks to validate the evaluation. The rates at which the model correctly assessed the responses are 89.6\% for video (last frame) and 97.7\% for audio (transcribed answer), showing a relatively high level of consistency.

\section{Text-Centric: Multimodal Reasoning Cases}
\label{app:multimodal_reasoning_cases}

We supplement the cases of Sora-2's solving multimodal reasoning questions of the text-centric tasks, as illustrated in Figure~\ref{fig:multimodal}.

\section{Supplementary Analysis and Results}

This section provides additional analyses and experimental results that complement the main findings. We present details on data leakage analysis, output modality experiments, reasoning process categorization, and manual evaluation results.

\subsection{Data Leakage Analysis}
\label{app:data_leakage_prompt}

As mentioned in Section~\ref{sec:exp_leakage}, we create new math evaluation problems to investigate potential data leakage as the reason for Sora-2's strong performance on text-centric tasks.

For each problem that we sampled from GSM8K~\citep{cobbe2021gsm8k} and MATH-500~\citep{cobbe2021training}, we used an LLM to derive a similar problem \textbf{with different numerical values} and possibly different contextual details while maintaining the overall difficulty. Qwen3-235B-A22B-Thinking~\citep{yang2025qwen3} and Gemini 2.5 Pro~\citep{comanici2025gemini} are used to derive the GSM8K problems the MATH-500 problems, respectively, with the prompts shown below.

\begin{table*}[t]
    \centering
    \small
    \setlength{\tabcolsep}{6pt}
    \caption{Accuracy (\%) on the visual puzzle tasks. * represents that \textbf{multiple-choice options} are provided for the \textbf{VLMs} due to evaluation need, as detailed in Appendix~\ref{app:visual_puzzle_eval}. Sora-2 is not provided with multiple-choice options across all 10 tasks.}
    \begin{tabularx}{\textwidth}{lCCCC}
        \toprule
        \multirow{2}{*}{\textbf{Task}} & \multirow{2}{*}{\textbf{Sora-2}} & \textbf{Gemini} & \textbf{GPT-5} & \textbf{Claude} \\
         & & \textbf{2.5 Pro} & \textbf{high} & \textbf{Sonnet 4.5} \\
        \midrule
        \multicolumn{5}{c}{\textit{Symmetry Tasks}} \\
        Hexagon Color Pattern Match. & 96.0 & 98.0 & 100.0 & 92.0 \\
        Grid Color Pattern Match. & 94.0 & 94.0 & 100.0 & 100.0 \\
        Grid Size Pattern Match.* & 85.4 & 87.5 & 95.8 & 62.5 \\
        Reflection Recognition \& Application* & 52.0 & 100.0 & 98.0 & 66.0 \\
        \textbf{Average} & 81.9 & 94.9 & 98.5 & 80.1 \\
        \midrule
        \multicolumn{5}{c}{\textit{Gradient Tasks}} \\
        Color Gradient Perception \& Application* & 45.8 & 83.3 & 35.4 & 93.8 \\
        Cycle Size Pattern Match.* & 58.0 & 84.0 & 98.0 & 46.0 \\
        \textbf{Average} & 51.9 & 83.7 & 66.7 & 69.9 \\
        \midrule
        \multicolumn{5}{c}{\textit{Compositionality Tasks}} \\
        Color Mixing Perception \& Application & 56.0 & 56.0 & 100.0 & 86.0 \\
        Shape Color Pattern Match. & 66.0 & 54.0 & 82.0 & 88.0 \\
        Rectangle Height Color Match. & 44.0 & 58.0 & 60.0 & 54.0 \\
        Grid Shape \& Size Pattern Match.* & 64.0 & 100.0 & 98.0 & 100.0 \\
        \textbf{Average} & 57.5 & 67.0 & 85.0 & 82.0 \\
        \midrule
        \textbf{Overall Average} & 66.2 & 81.5 & 86.8 & 78.8 \\
        \bottomrule
    \end{tabularx}
    \label{tab:visual_puzzle_result}
\end{table*}

\begin{promptbox}{Prompt for adapting the GSM8K problems}
\textit{Given a grade school math problem and its solution, derive a new problem that is similar in the underlying problem-solving structure but with different numbers and, if possible, with different context and way of expression.\\
Ensure the new problem is solvable with an integer answer and maintains the same level of difficulty.
Provide the solution to the derived problem using the same style and format as the original.
Enclose your problem in \textless problem\textgreater and \textless /problem\textgreater tags, and your solution in \textless solution\textgreater and \textless /solution\textgreater tags.\\
Original Problem and Solution:\\
Problem: \{original\_problem\}\\
Solution: \{original\_solution\}}
\end{promptbox}

\begin{promptbox}{Prompt for adapting the MATH-500 problems}
\textit{Given a math problem, derive a new problem that is similar in the underlying problem-solving structure but with different numbers and, if possible, with different context and way of expression.\\
Ensure the new problem is solvable and the complexity of its final answer is also similar to the original answer. For example, if the original answer is a simple integer (without any need to round), the new answer should also be a simple integer (also without any need to round).\\
Maintain the same level of difficulty.\\
Carefully analyze the original problem, think about the underlying structure, and carefully design the new problem to meet all the requirements above.\\
Provide the derived problem and a detailed solution to this new problem.\\
Enclose your problem in \textless problem\textgreater and \textless /problem\textgreater tags, and your solution in \textless solution\textgreater and \textless /solution\textgreater tags, and the final answer in \textless answer\textgreater and \textless /answer\textgreater tags.\\
Original Problem and Answer:\\
Problem: \{original\_problem\}\\
Answer: \{original\_answer\}}
\end{promptbox}

\begin{figure*}[h]
\centering 
\includegraphics[scale=0.5]{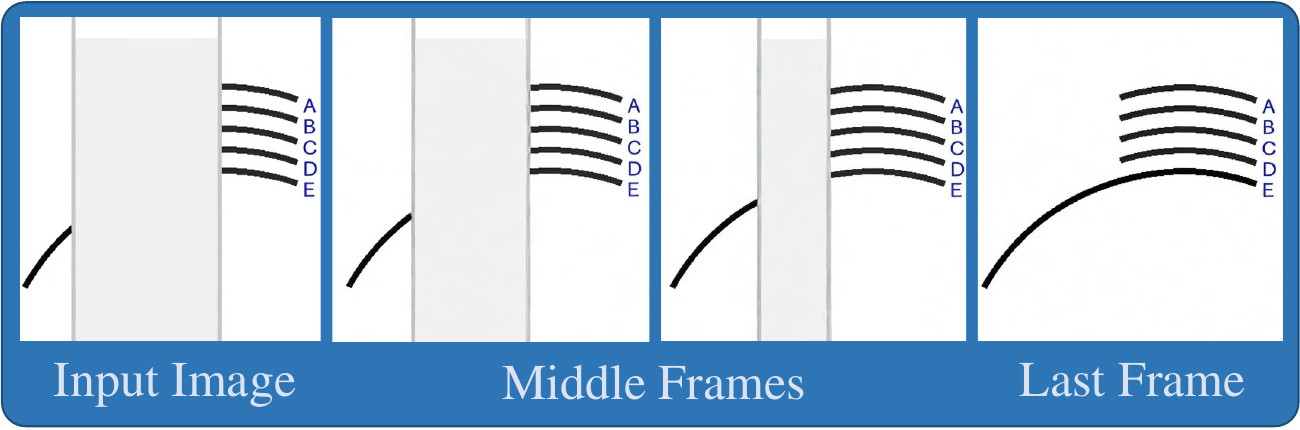}
\caption{\textbf{Sora-2 solving an Arc Connect puzzle.} Prompt: \textit{"One arc on the left continues across the masked band to one of the arcs on the right. Which labeled arc matches? Remove the masked band quickly while keeping the arcs still. Speak out the answer in phonetic alphabet. In portrait. Static Camera. No zoom."} Sora-2 successfully removes the band. Details: Section~\ref{sec:output_form}.}
\label{fig:case_arc}
\end{figure*}

\subsection{Output Modality Analysis}
\label{sec:output_form}

To explore how output form affects Sora-2 performance, we designed the Arc Connect puzzle, which requires determining which right arc connects to the left arc to form part of a circle. An example of Sora-2 solving an Arc Connect puzzle is shown in Figure~\ref{fig:case_arc}. 
The evaluation methods of Sora-2 on Arc Connect puzzle are defined as follows:
\begin{itemize}
    \item \textbf{Audio Option:} The prompt instructs model to speak out the option. Audio is extracted from generated video and transcribed to find the audio option.
    
    \item \textbf{Last Frame Option:} Last frame is extracted from the video. An evaluation program checks which right arc is connected to the left arc. If only one right arc is connected, the option is that option letter (``A'' to ``E'').
    
    \item \textbf{Major Frame Option:} For every 5 frames in the video, one frame is extracted and fed to the evaluation program, getting option of this frame. Major Frame Option is the majority vote result of all chosen frames.
\end{itemize}

\subsection{Prompt Rewriting in Wan 2.5: Case Study}
\label{app:wan_prompt_rewrite_case}

As discussed in Section~\ref{sec:source_text_reasoning_ability}, Wan 2.5's text-centric reasoning ability is almost entirely attributed to its prompt rewriter model. Here we provide a concrete example demonstrating how prompt rewriting transforms a reasoning task into explicit step-by-step visual instructions for the video generation component.

\subsubsection{Example: GSM8K Problem}
\label{app:wan_prompt_rewrite_problem}

\begin{figure*}[t]
    \centering
    \begin{subfigure}[t]{0.48\textwidth}
        \centering
        \includegraphics[width=\linewidth]{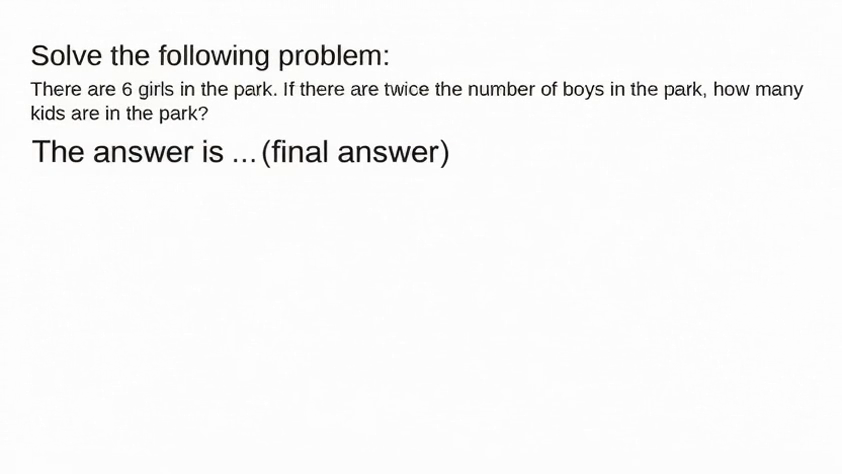}
        \caption{Without prompt rewriting (\texttt{prompt\_extend=false}): The model generates meaningless or incorrect content.}
        \label{fig:wan_no_rewrite}
    \end{subfigure}
    \hfill
    \begin{subfigure}[t]{0.48\textwidth}
        \centering
        \includegraphics[width=\linewidth]{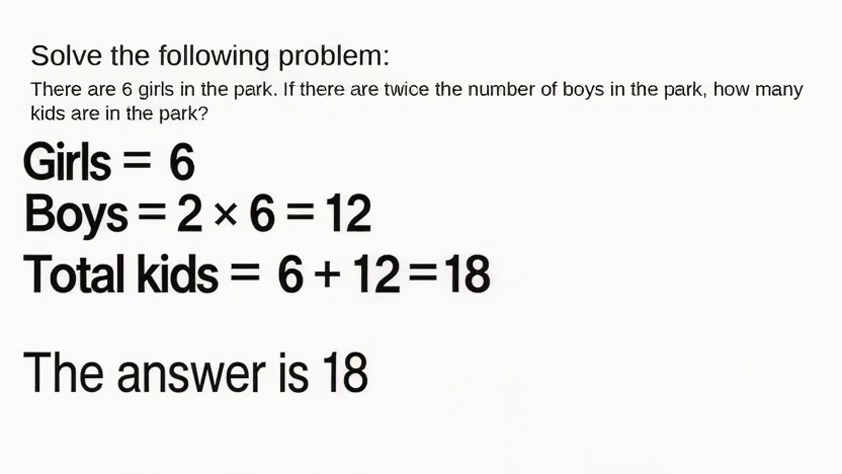}
        \caption{With prompt rewriting (\texttt{prompt\_extend=true}): The model correctly displays solution steps specified in the rewritten prompt (Appendix~\ref{app:wan_prompt_rewrite_problem}).}
        \label{fig:wan_with_rewrite}
    \end{subfigure}
    \caption{Visual comparison of Wan 2.5's outputs with and without prompt rewriting on the same GSM8K problem. The dramatic difference demonstrates that the reasoning capability resides in the prompt rewriter rather than the video generation model itself.}
    \label{fig:wan_prompt_rewrite_comparison}
\end{figure*}

\begin{table}[t]
    \centering
    
    \begin{minipage}[t]{0.49\textwidth}
        \centering
        \small
        \setlength{\tabcolsep}{6pt}
        \caption{Accuracy (\%) on the ARC-AGI-2 task. We display each sample in an image and send it to Sora-2 and VLMs. Details: Section~\ref{sec:arc_agi_2}.}
        \label{tab:arcagi_leaderboard_transposed} 
        \resizebox{\linewidth}{!}{%
        \begin{tabular}{lccccc}
            \toprule
            \multirow{2}{*}{\textbf{Task}} & \multirow{2}{*}{\textbf{Sora-2}} & \textbf{Gemini} & \textbf{GPT-5} & \textbf{Claude} & \multirow{2}{*}{\textbf{Grok 4}} \\
             & & \textbf{2.5 Pro} & \textbf{high} & \textbf{Sonnet 4.5} &\\
            \midrule
            ARC-AGI-2 & 1.3 & 1.9 & 0.5 & 5.3 &2.7\\
            \bottomrule
        \end{tabular}
        }
    \end{minipage}
    \hfill%
    \begin{minipage}[t]{0.49\textwidth}
        \centering
        \small
        \setlength{\tabcolsep}{2pt}
        \caption{Manual evaluation of 100 randomly chosen ARC-AGI-2 samples.}
        \vspace{2pt}
        \resizebox{\linewidth}{!}{%
        \begin{tabular}{lccccccc}
            \toprule
            \multirow{2}{*}{\textbf{Category}} & \textbf{Fully} & \textbf{Mostly} & \textbf{Partially} &
            \multicolumn{4}{c}{\textbf{Wrong}} \\
            \cmidrule(lr){5-8}
             & \textbf{Correct}& \textbf{Correct} & \textbf{Correct} & & Did Nothing & Others & \\
            \midrule
            Count & 3.0 & 14.0 & 28.0 & & 42.0 & 13.0 & \\
            \bottomrule
        \end{tabular}
        }
        \label{tab:arcagi_manual}
    \end{minipage}%
    
\end{table}

The rewritten prompt explicitly specifies the solution steps and visual elements to be generated, effectively solving the problem before video generation. This transformation explains why disabling the prompt rewriter leads to nearly zero accuracy (Table~\ref{tab:wan2.5_text_reason_test}).

\begin{promptbox}{Original Prompt (Without Explicit Rewriting)}
\textit{Solve the problem step by step on the given whiteboard. Give the final answer by writing ``The answer is ... (final answer)''. No oral explanation was provided during the written process of solving the problem, but the final answer was stated orally in the end.}

\textit{Problem: There are 6 girls in the park. If there are twice the number of boys in the park, how many kids are in the park?}
\end{promptbox}

\begin{promptbox}{Rewritten Prompt (After Prompt Rewriting)}
\textit{The problem is presented on a whiteboard. A hand writes `Girls = 6'. Then, `Boys = 2 × 6 = 12' appears. Next, `Total kids = 6 + 12 = 18' is written. Finally, `The answer is 18' is written on the board. A voice states: `The answer is 18'.}
\end{promptbox}

\subsubsection{Visual Comparison}

Figure~\ref{fig:wan_prompt_rewrite_comparison} shows the visual outputs generated by Wan 2.5 with and without prompt rewriting enabled.

\subsection{Evaluation Results of ARC-AGI-2}
We show the results in Table~\ref{tab:arcagi_leaderboard_transposed}.

\subsection{Manual Evaluation of ARC-AGI-2}
\label{sec:manual_eval}

To provide a more fine-grained assessment of Sora-2's performance on ARC-AGI-2 beyond binary correctness, we manually evaluated 100 randomly selected samples and categorized them into different quality levels. The results are in Table~\ref{tab:arcagi_manual}.

\subsection{Reasoning Process Categorization}
\label{app:solution_process_categories}

\begin{figure}[htbp]
    \begin{minipage}[t]{0.55\textwidth}
    \vspace{0pt} 
    In Section~\ref{sec:text_reason_process}, we analyzed Sora-2's reasoning processes for text-centric tasks. Here we provide the detailed categorization scheme. We categorize the solution process into five categories:
    \vspace{3pt}
    \begin{enumerate}[leftmargin=*]
        \item \textbf{Completely Correct:} The solution has a clear and correct process without any errors.
        \vspace{1pt}
        \item \textbf{Logic Correct with Writing Errors:} The solution contains expressional mistakes, but the overall logic is identifiable and correct.
        \vspace{1pt}
        \item \textbf{Unreadable or Incorrect Logic:} The writing is too disorganized or contains too many errors to discern the reasoning, or it exhibits clear logical mistakes or major omissions.
        \vspace{1pt}
        \item \textbf{Missing Solution Process:} Necessary steps are absent; apart from the final answer, the response is blank or contains only meaningless scribbles (i.e., lines, circles, etc).
        \vspace{1pt}
        \item \textbf{Process Unnecessary:} The problem itself does not require a written process to solve.
    \end{enumerate}
    \vspace{3pt}
    Distribution of these categories is illustrated in Figure~\ref{fig:reason_process_pie_chart}.
    \end{minipage}
    \hfill
    \begin{minipage}[t]{0.4\textwidth}
        \vspace{4pt} 
        \centering
        \includegraphics[width=\linewidth]{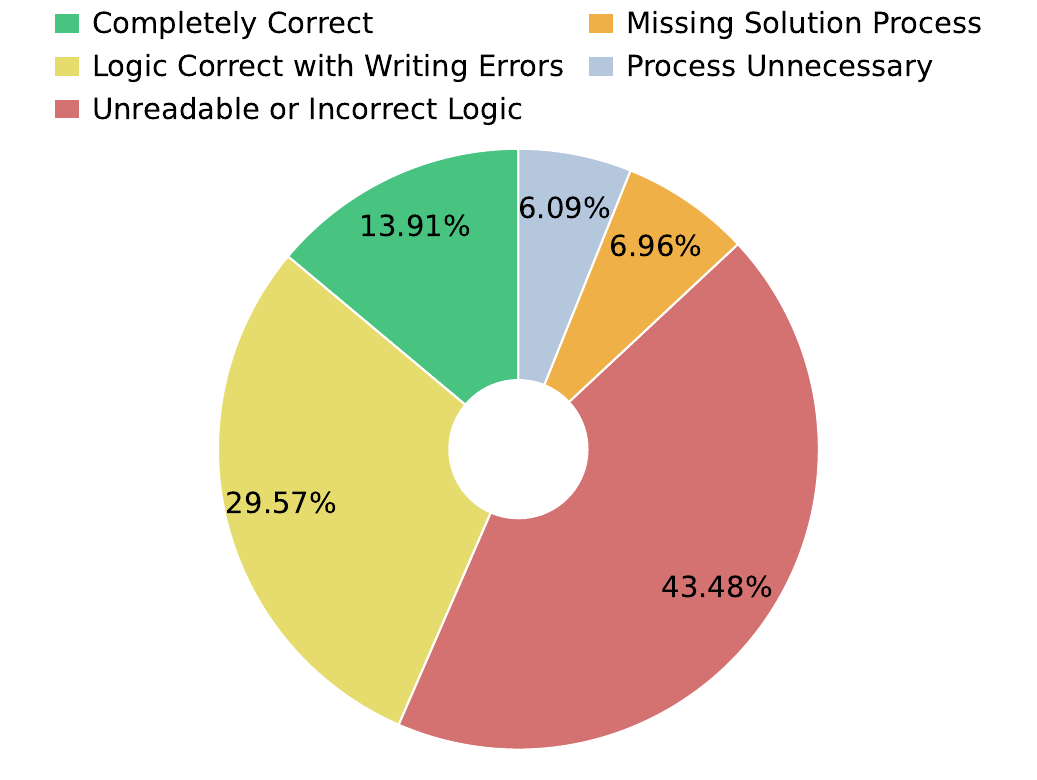}
        \vspace{4pt}
        \caption{Distribution of reasoning process categories for correct answers. The prevalence of ``Unreadable or Incorrect Logic'' (43.48\%) highlights Sora-2's difficulty in generating coherent reasoning.}
        \label{fig:reason_process_pie_chart}
    \end{minipage}
\end{figure}

\section{Prompts}
In this section, we list all the prompts used for evaluation in detail.

\subsection{Vision-Centric Tasks}
\subsubsection{Eyeballing Puzzles}
We list prompts used for each puzzle as below.
\begin{promptbox}{Prompt for Arc Connect}
\textbf{Sora-2 prompt:}
\textit{One arc on the left continues across the masked band to one of the arcs on the right. Which labeled arc matches? Remove the masked band quickly while keeping the arcs still. Speak out the answer in phonetic alphabet. In portrait. Static Camera. No zoom.}

\textbf{VLMs prompt:}
\textit{One arc on the left continues across the masked band to one of the arcs on the right. Which labeled arc matches? Answer an option in A-E.}
\end{promptbox}

\begin{promptbox}{Prompt for Angle Bisector}
\textbf{Sora-2 prompt:}
\textit{Draw a black line bisecting the angle. Speak out which option is on the bisector using phonetic alphabet and mark that red. In portrait, no zoom, no pan.}

\textbf{VLMs prompt:}
\textit{Which option is on the bisector of the angle? Answer an option in A-E.}
\end{promptbox}

\begin{promptbox}{Prompt for Circle Center}
\textbf{Sora-2 prompt:}
\textit{Mark the center of the circle red. Speak out which option is the center using phonetic alphabet. In portrait, no zoom, no pan.}

\textbf{VLMs prompt:}
\textit{Which option is the center of the circle? Answer an option in A-E.}
\end{promptbox}

\begin{promptbox}{Prompt for Circle Tangent Line}
\textbf{Sora-2 prompt:}
\textit{Draw a black line tangent to the circle at the highlighted point. Speak out which option lies on this tangent line in phonetic alphabet and mark that red. In portrait, no zoom, no pan.}

\textbf{VLMs prompt:}
\textit{Which option lies on the line that is tangent to the circle at the highlighted point? Answer an option in A-E.}
\end{promptbox}

\begin{promptbox}{Prompt for Circle Tangent Point}
\textbf{Sora-2 prompt:}
\textit{Draw the tangent line from the external point to the circle in black. Paint the point of tangency red. Speak out which option is the point using phonetic alphabet. In portrait, no zoom, no pan.}

\textbf{VLMs prompt:}
\textit{Which option is the tangent point on the circle from the external point? Answer an option in A-E.}
\end{promptbox}

\begin{promptbox}{Prompt for Circumcenter}
\textbf{Sora-2 prompt:}
\textit{Mark the circumcenter of the triangle red. Speak out which option is the circumcenter using phonetic alphabet. In portrait, no zoom, no pan.}

\textbf{VLMs prompt:}
\textit{Which option is the circumcenter of the triangle? Answer an option in A-E.}
\end{promptbox}

\begin{promptbox}{Prompt for Fermat Point}
\textbf{Sora-2 prompt:}
\textit{Find the Fermat point of the triangle. Mark the point red. Speak out which option is the Fermat point using the phonetic alphabet. In portrait, no zoom, no pan.}

\textbf{VLMs prompt:}
\textit{Which option is the Fermat point of the triangle? Answer an option in A-E.}
\end{promptbox}

\begin{promptbox}{Prompt for Incenter}
\textbf{Sora-2 prompt:}
\textit{Mark the incenter of the triangle red. Speak out which option is the incenter using phonetic alphabet. In portrait, no zoom, no pan.}

\textbf{VLMs prompt:}
\textit{Which option is the incenter of the triangle? Answer an option in A-E.}
\end{promptbox}

\begin{promptbox}{Prompt for Isosceles Trapezoid}
\textbf{Sora-2 prompt:}
\textit{Find the fourth vertex that completes the isosceles trapezoid. Mark the fourth vertex red. Speak out which option is the fourth vertex using phonetic alphabet. In portrait, no zoom, no pan.}

\textbf{VLMs prompt:}
\textit{Which option is the fourth vertex of the isosceles trapezoid? Answer an option in A-E.}
\end{promptbox}

\begin{promptbox}{Prompt for Midpoint}
\textbf{Sora-2 prompt:}
\textit{Connect the two large circles and mark the midpoint as red. Speak out which option is the midpoint using phonetics alphabet. In portrait, no zoom, no pan.}

\textbf{VLMs prompt:}
\textit{Which option is the midpoint of the two circles? Answer an option in A-E.}
\end{promptbox}

\begin{promptbox}{Prompt for Orthocenter}
\textbf{Sora-2 prompt:}
\textit{Find the orthocenter (intersection of altitudes) of the triangle and mark it red. Speak out which option is the orthocenter using phonetic alphabet. In portrait, no zoom, no pan.}

\textbf{VLMs prompt:}
\textit{Which option is the orthocenter of the triangle? Answer an option in A-E.}
\end{promptbox}

\begin{promptbox}{Prompt for Parallel}
\textbf{Sora-2 prompt:}
\textit{Draw a black line through the small circle and parallel to the existing line. Speak out which option is on the new line using phonetic alphabet and mark that red. In portrait, no zoom, no pan.}

\textbf{VLMs prompt:}
\textit{Draw a line through the small circle and parallel to the existing line, which option is on it? Answer an option in A-E.}
\end{promptbox}

\begin{promptbox}{Prompt for Parallelogram}
\textbf{Sora-2 prompt:}
\textit{Draw a black parallelogram with two sides given. Mark the fourth vertex red. Speak out which option is the fourth vertex using phonetics alphabet. In portrait, no zoom, no pan.}

\textbf{VLMs prompt:}
\textit{Which option is the fourth vertex of the parallelogram with two sides given? Answer an option in A-E.}
\end{promptbox}

\begin{promptbox}{Prompt for Perpendicular}
\textbf{Sora-2 prompt:}
\textit{Draw a black line perpendicular to the existing line and passing the small circle. Speak out which option is on the line using phonetic alphabet and mark that red. In portrait, no zoom, no pan.}

\textbf{VLMs prompt:}
\textit{Which option is on the line perpendicular to the black line and passing the small circle? Answer an option in A-E.}
\end{promptbox}

\begin{promptbox}{Prompt for Perpendicular Bisector}
\textbf{Sora-2 prompt:}
\textit{Draw a black line that is the perpendicular bisector of the segment between the two small circles. Speak out which option is on the line using phonetic alphabet and mark that red. In portrait, no zoom, no pan.}

\textbf{VLMs prompt:}
\textit{Which option is on the perpendicular bisector of the segment connecting the two small circles? Answer an option in A-E.}
\end{promptbox}

\begin{promptbox}{Prompt for Ray Intersection}
\textbf{Sora-2 prompt:}
\textit{Extend the three black lines and mark the intersection point as red. Speak out which option is the intersection point using phonetics alphabet. In portrait, no zoom, no pan.}

\textbf{VLMs prompt:}
\textit{Which option is the intersection point of the three lines? Answer an option in A-E.}
\end{promptbox}

\begin{promptbox}{Prompt for Ray Reflection}
\textbf{Sora-2 prompt:}
\textit{Draw the ray of light starting from the small circle and reflecting off the line in black. Speak out which option the reflected ray will pass through using phonetic alphabet and mark it red. In portrait, no zoom, no pan.}

\textbf{VLMs prompt:}
\textit{A ray of light starts from the small circle and reflects off the line. Which option will the reflected ray pass through? Answer an option in A-E.}
\end{promptbox}

\begin{promptbox}{Prompt for Point Reflection}
\textbf{Sora-2 prompt:}
\textit{Reflect the small circle across the line. Mark the reflection red and speak out which option is the reflected point using phonetic alphabet. In portrait, no zoom, no pan.}

\textbf{VLMs prompt:}
\textit{Which option is the reflection of the small circle across the line? Answer an option in A-E.}
\end{promptbox}

\begin{promptbox}{Prompt for Right Triangle}
\textbf{Sora-2 prompt:}
\textit{Out of the 5 points, 3 form a right-angled triangle. Mark the vertex with the right angle in red. Speak out which option is the right-angle vertex using phonetic alphabet. In portrait, no zoom, no pan.}

\textbf{VLMs prompt:}
\textit{Which option is the vertex of the right angle, given that exactly three of the five options form a right-angled triangle? Answer an option in A-E.}
\end{promptbox}

\begin{promptbox}{Prompt for Square Outlier}
\textbf{Sora-2 prompt:}
\textit{Four of the five options form a square. Mark the fifth point red. Speak out which option is the fifth point using phonetics alphabet. In portrait, no zoom, no pan.}

\textbf{VLMs prompt:}
\textit{Four of the five options form a square. Which option is the fifth point? Answer an option in A-E.}
\end{promptbox}

\begin{promptbox}{Prompt for Triangle Center}
\textbf{Sora-2 prompt:}
\textit{Mark the center of the triangle red. Speak out which option is the center using phonetic alphabet. In portrait, no zoom, no pan.}

\textbf{VLMs prompt:}
\textit{Which option is the center of the triangle? Answer an option in A-E.}
\end{promptbox}

\subsubsection{Visual Puzzles}

\label{app:visual_puzzle_prompts}
\begin{promptbox}{Prompt for Tasks 1, 2, 7, 8 and 9}
\textbf{Sora-2 prompt:} \textit{What is the missing color of the part denoted with a question mark? This part should be completely filled with the correct color while the other parts should be unchanged. The question mark disappears. Then nothing happens and the scene remains static. Do not zoom in or out, or change the positions of the shapes.}

\textbf{VLMs prompt:} \textit{What is the missing color of the part denoted with a question mark?}
\end{promptbox}

\begin{promptbox}{Prompt for Task 5 (Color Gradient Perception \& Application)}

\textbf{Sora-2 prompt:} \textit{What is the missing color of the part denoted with a question mark? This part should be completely filled with the correct color (not white or the original grey) to match the pattern in the image while the other parts should be unchanged. The question mark disappears. Then nothing happens and the scene remains static. Do not zoom in or out, or change the positions of the shapes.}

\textbf{VLMs prompt:} \textit{What is the missing color of the part denoted with a question mark? Options: ...} (Four options.)
\end{promptbox}

\begin{promptbox}{Prompt for Task 3 (Grid Size Pattern Matching) and 6 (Cycle Size Pattern Matching)}
\textbf{Sora-2 prompt:} \textit{What is the size of the missing part denoted with a question mark? This part should be replaced with the correct circle while the other circles should be unchanged. The question mark disappears. Then nothing happens and the scene remains static. Do not zoom in or out, or change the positions of the shapes.}

\textbf{VLMs prompt:} \textit{What is the size of the missing circle denoted with a question mark? Options: small, medium, large} (The three options are randomly shuffled.)
\end{promptbox}

\begin{promptbox}{Prompt for Task 10 (Grid Shape \& Size Pattern Matching)}
\textbf{Sora-2 prompt:} \textit{What is the size of the missing part denoted by a question mark? This part should be replaced with the correct shape while the other shapes should be unchanged. The question mark disappears. Then nothing happens and the scene remains static. Do not zoom in or out, or change the positions of the shapes.}

\textbf{VLMs prompt:} \textit{What is the size of the missing part denoted by a question mark? Options: small, medium, large.} (The three options are randomly shuffled.)
\end{promptbox}

\begin{promptbox}{Prompt for Task 4 (Reflection Recognition \& Application)}

\textbf{Sora-2 prompt:} \textit{What is the missing shape denoted by a question mark? The question mark area should be replaced with the correct shape while the other shapes should be unchanged. The question mark disappears. Then nothing happens and the scene remains static. Do not zoom in or out, or change the positions of the shapes.}

\textbf{VLMs prompt:} \textit{What is the missing shape denoted by a question mark? Options: triangle, square, pentagon, hexagon.} (The four options are randomly shuffled.)

\end{promptbox}
\subsection{Text-Centric Tasks}

\subsubsection{Generation}

\begin{promptbox}{Prompt for problems from GSM8K, MATH-500, AIME and GPQA-diamond}
\textit{Solve the problem step by step on the given whiteboard. No oral explanation was provided during the written process of solving the problem, but the final answer was stated orally in the end, which is also clearly written.\\
Problem: \{problem\}}
\end{promptbox}

\begin{promptbox}{Prompt for problems from BBH, MMLU, MMLU-Pro and SuperGPQA-easy}
\textit{A short video explaining a multiple-choice question.}

\textit{**Visual Setup:**\\
- **Background:** A solid, pure white background throughout the entire video.\\
- **Layout:** Split-screen layout.\\
    - Clearly displays the question and multiple-choice options. Use a large, clean, and easy-to-read font.\\
    - No presenter or other irrelevant content to the question.\\
    - The question is displayed at the top center with ``Question:'' as the title\\
    - Multiple-choice options (A, B, C, etc.) are listed below
    - A ``Correct Answer: \_\_\_\_\_'' line appears at the bottom with appropriate spacing from the edge\\
    - All text uses clear, easy-to-read fonts\\
**Content to Display:**\\
Question:\\
\{question\}\\
Correct Answer: \_\_\_\_\_ (fill this in after explanation)\\
**Requirements:**\\
- Directly state the correct answer through audio narration (e.g., ``The correct answer is A'' or ``The answer is True'')\\
- Fill in the correct answer in the ``Correct Answer: \_\_\_\_\_'' line on screen\\
- No need for explanation or reasoning - just clearly announce the answer\\
**Style \& Tone:** Clear and articulate voice, professional tone, direct and concise.}
\end{promptbox}

\begin{promptbox}{Prompt for problems from MathVista, MathVision, MMBench and MMMU}
    \textit{Question: \{question\}\\
    Generate a video showing the solution process}
\end{promptbox}
\subsubsection{Evaluation}
\label{app:eval_prompt_text-centric}

For the text-centric tasks, we use GPT-4o~\citep{openai2024gpt4o} as the judge model to evaluate the answer from the video and the audio independently, with the prompts shown below. For audio transcription, we use OpenAI's whisper model (whisper-1) via its API.

\begin{promptbox}{Prompt for Evaluating the Answer from the Video}

\textbf{System prompt:}

\textit{You are an expert answer checker for educational videos. Your task is to determine if an image (the last frame of a solution video) displays the correct answer to a given question.}

\textit{Rules:\\
0. First, determine the visible answer from the image using this priority:\\
   - If there is an explicit statement indicating the answer (e.g., ``The answer is ...''), use that answer.\\
   - Else, check for an answer marked by a symbol such as box, circle, underline, arrow, etc. If multiple positions are marked but show different results, respond 'no' immediately.\\
   - Else, use the bottom-rightmost result in the image as the visible answer.\\
1. Compare the visible answer in the image with the provided correct answer\\
2. Be strict but reasonable - minor formatting differences are acceptable if the core answer is correct\\
3. For multiple choice questions, check if the correct option (A, B, C, etc.) is clearly marked or highlighted\\
4. For numerical answers, check if the number matches (ignore minor formatting like ``4'' vs ``4.0'')\\
5. For text answers, check if the key content matches (ignore case sensitivity and minor punctuation)\\
6. You must respond with ONLY 'yes' or 'no', nothing else}

\textbf{User instruction prompt: }

\textit{Question: \{question\}}

\textit{Correct answer: \{correct\_answer\}}

\textit{Does the image show the correct answer?}

(The last frame of the generated video is also provided for the model.)

\end{promptbox}

\begin{promptbox}{Prompt for Evaluating the Answer from the Audio}

\textbf{System prompt:}

\textit{You are an expert answer checker for educational video transcripts. Your task is to determine if an audio transcript from a solution video contains the correct answer to a given question.}

\textit{Rules:\\
1. Check if the transcript explicitly states or clearly implies the correct answer\\
2. Be lenient with phrasing - the transcript may explain the answer in different words\\
3. For multiple choice questions, check if the correct option (A, B, C, etc.) is mentioned\\
4. For numerical answers, check if the number is stated (ignore surrounding explanation)\\
5. For text answers, check if the key concept is explained correctly\\
6. Common phrases like ``the correct answer is...'', ``the answer is...'', ``it should be...'' indicate the answer\\
7. You must respond with ONLY 'yes' or 'no', nothing else}

\textbf{User instruction prompt: }

\textit{Question: \{question\}}

\textit{Correct answer: \{correct\_answer\}}

\textit{Audio transcript: \{transcript\}}

\textit{Does the transcript provide the correct answer?}

\end{promptbox}


\end{document}